\documentclass[10pt,twocolumn,letterpaper]{article}

\usepackage{cvpr}
\usepackage[utf8]{inputenc}
\usepackage{graphicx}
\usepackage{color}
\usepackage{times}
\usepackage{amsmath,mathtools}
\usepackage{amsfonts}
\usepackage{amsthm}	
\usepackage{amssymb}
\usepackage{enumerate}
\usepackage{algorithm} 
\usepackage{algpseudocode}
\usepackage{booktabs}
\usepackage{tabularx}
\usepackage{array}
\usepackage{bm}
\usepackage{url}
\usepackage{bbm}
\usepackage{enumitem}
\usepackage{stackengine}
\usepackage[algo2e,inoutnumbered,linesnumbered,algoruled,vlined]{algorithm2e}
\usepackage{xcolor}
\usepackage{multirow}
\usepackage{authblk}
\usepackage{pifont}
\usepackage[symbol*]{footmisc}

\usepackage[caption=false]{subfig}
\usepackage[breaklinks=true,bookmarks=false]{hyperref}

\cvprfinalcopy % *** Uncomment this line for the final submission
\usepackage{cleveref}

\def\cvprPaperID{109} % *** Enter the CVPR Paper ID here

\ifcvprfinal\fi

\DefineFNsymbolsTM{otherfnsymbols}{%
  \textbullet \circ
  \textsection   \mathsection
  \textdagger    \dagger
  \textdaggerdbl \ddagger
  \textasteriskcentered *
  \textbardbl    \|%
  \textparagraph \mathparagraph
}%

\setfnsymbol{otherfnsymbols}
\newcommand{\cmark}{\ding{61}}%
\newcommand{\xmark}{*}%
\newcommand{\zmark}{\ding{67}}%

\newcommand{\x}{\mathbf{x}}

\newcommand{\Pm}{\mathbf{P}}
\newcommand{\Rot}{\mathbf{R}}

\newcommand{\X}{\mathbf{X}}

\newcommand{\q}{\mathbf{q}}

\newcommand{\p}{\mathbf{p}}
\newcommand{\tb}{\mathbf{t}} 
\newcommand{\fb}{\mathbf{f}} 
\newcommand{\T}{\mathbf{T}}

\newcommand{\F}{\mathbf{F}}

\newcommand{\R}{\mathbb{R}}

\crefname{section}{§}{§§}
\Crefname{section}{§}{§§}

\crefname{section}{§}{§§}
\Crefname{section}{§}{§§}
\crefname{thm}{Thm.}{Thm}
\crefname{eq}{Eq.}{Eq}
\crefname{figure}{Fig.}{Figure}
\crefname{dfn}{Dfn.}{Dfn}
\crefname{table}{Tab.}{Table}

\newcommand{\insertimageC}[5]{ % scale, filename, caption, label, location
\begin{figure}[#5]
\centering
\includegraphics[width=#1\linewidth, clip=true]{figures/#2}
\caption{#3}
\label{#4}
\end{figure}
}

\newcommand{\insertimageStar}[5]{ % scale, filename, caption, label, location
\begin{figure*}[#5]
\centering
\includegraphics[width=#1\linewidth, clip=true]{figures/#2}
\caption{#3}
\label{#4}
\end{figure*}
}

\algnewcommand\algorithmicinput{\textbf{Input:}}
\algnewcommand\INPUT{\item[\algorithmicinput]}
\algnewcommand\algorithmicoutput{\textbf{Output:}}
\algnewcommand\OUTPUT{\item[\algorithmicinput]}

\newcommand{\comment}[1]{}

\newcommand{\suchthat}{\;\ifnum\currentgrouptype=16 \middle\fi|\;}

\begin{document}

%%%%%%%%% TITLE
\title{3D Local Features for Direct Pairwise Registration}

\author[ \cmark, \xmark, \zmark]{Haowen Deng}
\author[ \cmark, \xmark]{Tolga Birdal}
\author[ \cmark, \xmark]{Slobodan Ilic}
\affil[\cmark]{ Technische Universit\"at M\"{u}nchen, Germany,\quad \xmark~Siemens AG, M\"{u}nchen, Germany}
%\affil[\xmark]{ Siemens AG, M\"{u}nchen, Germany}
\affil[\zmark]{ National University of Defense Technology, China
%{\normalsize \tt{haowen.deng@tum.de\,}},  {\normalsize \tt{\,tolga.birdal@tum.de\,}}, {\normalsize \tt{\,slobodan.ilic@siemens.com}}
}

\maketitle
%\thispagestyle{empty}

%%%%%%%%% ABSTRACT
\begin{abstract}
We present a novel, data driven approach for solving the problem of registration of two point cloud scans. Our approach is direct in the sense that a single pair of corresponding local patches already provides the necessary transformation cue for the global registration.
To achieve that, we first endow the state of the art PPF-FoldNet~\cite{Deng_2018_ECCV} auto-encoder (AE) with a pose-variant sibling, where the discrepancy between the two leads to pose-specific descriptors. Based upon this, we introduce \textit{RelativeNet}, a relative pose estimation network to assign correspondence-specific orientations to the keypoints, eliminating any local reference frame computations. Finally, we devise a simple yet effective hypothesize-and-verify algorithm to quickly use the predictions and align two point sets. Our extensive quantitative and qualitative experiments suggests that our approach outperforms the state of the art in challenging real datasets of pairwise registration and that augmenting the keypoints with local pose information leads to better generalization and a dramatic speed-up.
\end{abstract}

% Learning local feature descriptors from 3D is an active and important research area. The capability of finding associations of local structures in point clouds simplifies various tasks such as correspondence computation, pose estimation, or retrieval. The de-facto standard first crafts/learns a local descriptor with an invariance flavor and then applies it to the tasks at hand. 
% In this paper, we argue that the tasks being tackled are critical for the capacity of the learned features and propose a novel deep point-cloud network for learning local descriptors tailored to direct 3D matching and registration.
%Our algorithm enjoys a weakly-supervised multi-task loss
%fusing unsupervised 3D reconstruction losses
%with self-supervised 3D pose and matching penalties.

%
\vspace{-2mm}\section{Introduction}
\label{sec:intro}
\insertimageC{1}{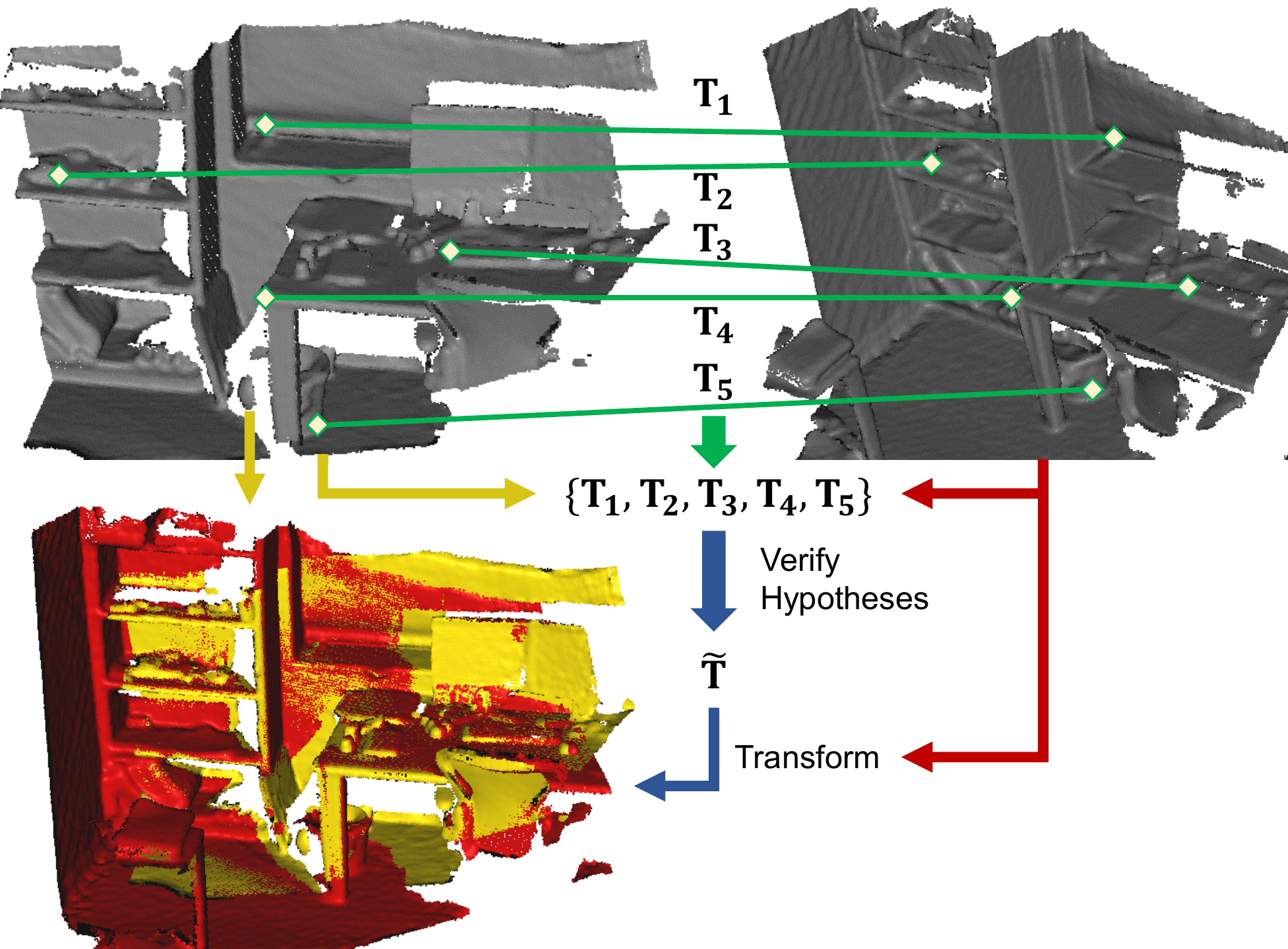}{Our method provides not only powerful features for establishing correspondences, but also directly predicts a rigid transformation attached to each correspondence. Final estimation of the rigid pose between fragment pairs can then be made efficiently by operating on the pool of pose predictions.\vspace{-4mm}}{fig:teaserDPR}{t!}
Learning and matching local features have fueled computer vision for many years. Scholars have first hand-crafted their descriptors~\cite{lowe2004distinctive} and with the advances in deep learning, devised data driven methods that are more reliable, robust and practical~\cite{krizhevsky2012imagenet,yi2016lift}. These developments in the image domain have quickly escalated to 3D where 3D descriptors~\cite{salti2014shot,zeng20163dmatch,deng2018ppfnet} have been developed. 

Having 3D local features at hand is usually seen as an intermediate step towards solving more challenging 3D vision problems. One of the most prominent of such problems is 3D pose estimation, where the six degree-of-freedom (6DoF) rigid transformations relating 3D data pairs are sought. This problem is also known as \textit{pairwise 3D registration}. While the quality of the intermediary descriptors is undoubtedly an important aspect towards good registration performance~\cite{guo2014performance}, directly solving the final problem at hand is certainly more critical. Unfortunately, contrary to 2D descriptors, the current deeply learned 3D descriptors~\cite{zeng20163dmatch,deng2018ppfnet,Deng_2018_ECCV} are still not tailored for the task we consider, i.e. they lack any kind of local orientation assignment and hence, any subsequent pose estimator is coerced to settle for nearest neighbor queries and exhaustive RANSAC iterations to robustly compute the aligning transformation. This is neither reliable nor computationally efficient.

In this paper, we argue that descriptors that are good for pairwise registration should also provide cues for direct computation of local rotations and propose a novel, robust and end-to-end algorithm for local feature based 3D registration of two point clouds (See~\cref{fig:teaserDPR}). We begin by augmenting the state-of-the-art unsupervised, 6DoF-invariant local descriptor PPF-FoldNet~\cite{Deng_2018_ECCV} with a deeply learned orientation. Via our pose-variant orientation learning, we can decouple the \textit{3D structure} from \textit{6DoF motion}. This can result in features solely explaining the pose variability up to a reasonable approximation. Our network architecture is shown in~\cref{fig:pipelineRelNet}. We then make the observation that locally good registration leads to good global alignment and vice versa. Based on that, we propose a simple yet effective \textit{hypothesize-and-verify} scheme to find the optimal alignment conditioned on an initial correspondence pool that is simply retrieved from the (mutually) closest nearest neighbors in the latent space. 
\insertimageStar{1}{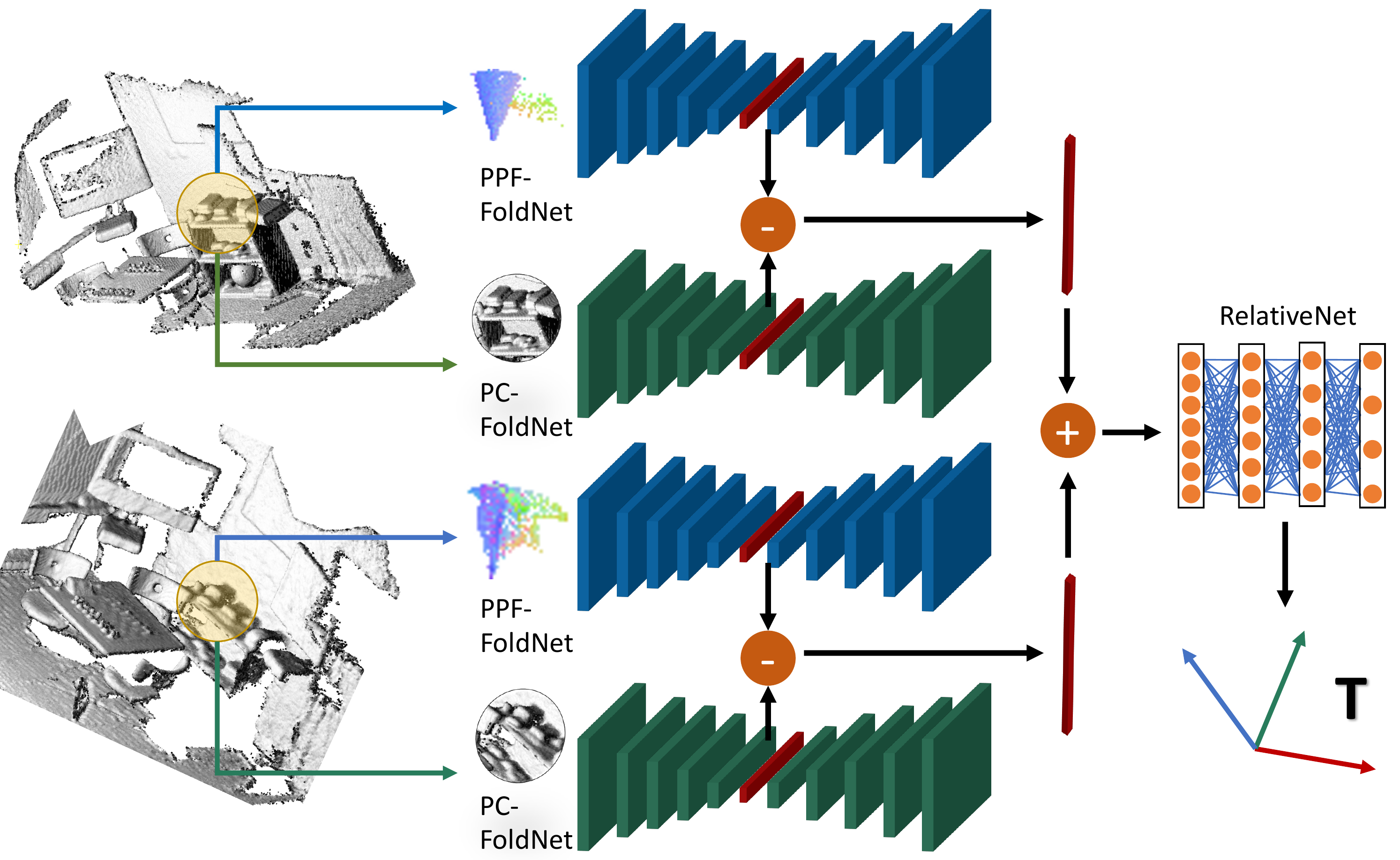}{Overview of proposed pipeline. Given two point clouds, we first feed all the patches into PPF-FoldNet and PC-FoldNet auto-encoders to extract invariant and pose-variant local descriptors, respectively. Patch pairs are then matched by their intermediate invariant features. The pairs that are found to match are further processed to compute the discrepancy between invariant PPF-based features and PC-based features. These ratio features belonging to pairs of matching keypoints are concatenated and sent into RelativeNet, generating relative pose predictions. Multiple signals are imposed on reconstruction, pose prediction and feature consistency during the training stage.\vspace{-3mm}}{fig:pipelineRelNet}{t!}

For the aforementioned idea to work well, the local orientations assigned to our keypoints (sampled with spatial uniformity) should be extremely reliable. Unfortunately, finding such repeatable orientations of local patches immediately calls for local reference frames (LRF), which are by themselves a large source of ambiguity and error~\cite{petrelli2011repeatability}. Therefore, we instead choose to learn to estimate \textit{relative} transformations instead of aiming to find a canonical frame. We find the relative motion to be way more robust and easier-to-train for than an LRF. To this end, we introduce \textit{RelativeNet}, a specialized architecture for relative pose estimation.

We train all of our networks end-to-end by combining three loss functions: 1) Chamfer reconstruction loss for the unsupervised PPF-FoldNet~\cite{Deng_2018_ECCV}, 2) Weakly-supervised relative pose cues for the transformation-variant local features, 3) A  feature-consistency loss which enforces the nearby points to give rise to nearby features in the embedding space. 
We evaluate our method extensively against multiple widely accepted benchmark datasets of 3DMatch-benchmark~\cite{zeng20163dmatch} and Redwood~\cite{Choi_2015_CVPR}, on the important tasks of feature matching and geometric registration. On our assessments, we improve the state of the art by $6.83\%$ in pairwise registration while reducing the runtime by 20 folds. This dramatic improvement in both aspects stems from the weak supervision making the local features capable of spilling rotation estimates and thereby easing the job of the final transformation estimator. The interaction of three multi-task losses in return enhances all predictions.
Overall, our contributions are:
\begin{enumerate}[itemsep=0ex]
\item Invariant + pose-variant network for local feature learning designed to generate pose-related descriptors that are insensitive to geometrical variations.
\item A multi-task training scheme which could assign orientations to matching pairs and simultaneously strengthen the learned descriptors for finding better correspondences.
\item Improvement of geometric registration performance on given correspondence set using direct network predictions both interms of speed and accuracy.

%as well as speedup of the procedure.
\end{enumerate}

\section{Related Work}
\label{sec:related_work}
\paragraph{Local descriptors}
There has been a long history of handcrafted features, designed by studying the geometric properties of local structures. FPFH~\cite{rusu2009fast}, SHOT~\cite{salti2014shot}, USC~\cite{tombari2010unique} and Spin Images~\cite{johnson1999using} all use different ideas to capture these properties. Unfortunately, the challenges of real data, such as the presence of noise, missing structures, occlusions or clutter significantly harm such descriptors~\cite{guo2014performance}. 
Recent trends in data driven approaches have encouraged the researchers to harness deep learning to surmount these nuisances. Representative works include 3DMatch~\cite{zeng20163dmatch}, PPFNet~\cite{deng2018ppfnet}, CGF~\cite{Khoury_2017_ICCV}, 3D-FeatNet~\cite{jian20183dfeat}, PPF-FoldNet~\cite{Deng_2018_ECCV} and 3D point-capsule networks~\cite{zhao20193d}, all outperforming the handcrafted alternatives by large margin. 
While the descriptors in 2D are typically complemented by the useful information of local orientation, derived from the local image appearance~\cite{lowe2004distinctive}, the nature of 3D data renders the task of finding a unique and consistent local coordinate frame way more challenging~\cite{Gojcic2019,petrelli2011repeatability}. Hence, none of the aforementioned works were able attach local orientation information to 3D patches. This motivates us to jointly consider descriptor extraction and the direct 3D alignment.
\vspace{-3mm}\paragraph{Pairwise registration} 
The approaches to pairwise registration fork into two main research directions. 

The first school tries to find an alignment of two point sets globally. Iterative closest point (ICP)~\cite{besl1992method} and its transcendents~\cite{toldo2010global,yang2013go,besl1992method,li20073d} alternatively hypothesize a correspondence set and minimize the 3D registration error optimizing for the rigid pose. Despite its success, making ICP outlier-robust is considered, even today, to be an open problem~\cite{Lawin_2018_CVPR,Eckart_2018_ECCV,vongkulbhisal2018inverse,bustos2018guaranteed}. Practical applications of ICP also incorporate geometric, photometric or temporal consistency cues~\cite{park2017colored} or odometry constraints~\cite{zhou2016fast}, whenever available. ICP is prone to the initialization and is known to tolerate only up to a $15-30^\circ$ misalignment~\cite{birdal2017cad,birdal2016online}. 

Another family branches off from Random Sample Consensus (RANSAC)~\cite{Fischler1981}. These works hypothesize a set of putative matches of keypoints and attempt to disable the erroneous ones via a subsequent rejection. The discovered inliers can then be used in a Kabsch-like~\cite{kabsch1976solution} algorithm to estimate the optimal transformation. A notable drawback of RANSAC is the huge amount of trials required, especially when the inlier ratio is low and the expected confidence of finding a correct subset of inliers is high~\cite{choi1997performance}. This encouraged the researchers to propose accelerations to the original framework, and at this time, the literature is filled with an abundance of RANSAC-derived methods~\cite{chum2008optimal,chum2003locally,korman2018latent,chum2005matching}, unified under the USAC framework~\cite{raguram2013usac}.

Even though RANSAC is now a well developed tool, heuristics associated to it facilitated the scholars to look for more \textit{direct} detection and pose estimation approaches, hopefully alleviating the flaws of feature extraction and randomized inlier maximization. Recently, the geometric hashing of point pair features (PPF)~\cite{birdal3dv2015,drost2010model,birdal2017point,hinterstoisser2016going,vidal20186d} is found to be the most reliable solution~\cite{bop}.
Another alternative includes 4-point congruent set (4-PCS)~\cite{aiger20084,bueno20184} further made efficient by the Super4PCS~\cite{mellado2014super} and generalized by~\cite{mohamad2015super}. 
As we will elaborate in the upcoming sections, our approach lies at the intersection of local feature learning and direct pairwise registration inheriting the good traits of both.
%Local features play an important role in many applications %seeking correspondences . 
%
%The discriminative power and robustness of local features %largely decide the performance of local correspondence %establishment, which further affect the performance of %higher-end applications like SLAM, 3D reconstruction and etc. %
\section{Method}
\label{sec:method}
%\subsection{Formulation}
Purely geometric local patches typically carry two pieces of information: (1) 3D \textit{structure}, summarized by the sample points themselves $\Pm = \{\p_i \,|\, \p_i \in \R^{N\times 3} \}$ where $\p=[x,y,z]^\top$ and (2) \textit{motion}, which in our context corresponds to the 3D transformation or the \textit{pose} $\T_i\in SE(3)$ holistically orienting and spatially positioning the point set $\Pm$:
\begin{equation}
\label{eq:se3}
SE(3) = \left \{
\mathbf{T} \in \R^{4\times 4} \colon \mathbf{T}=\begin{bmatrix} 
\Rot & \mathbf{t} \\ 
\mathbf{0}^\top & 1 
\end{bmatrix}
\right\}.
\end{equation}
where $\Rot \in SO(3) \text{ and } \mathbf{t} \in \R^3$.
A point set $\Pm_i$, representing a local patch is generally viewed as a transformed replica of its canonical version $\Pm^c_i$: $\Pm_i= \T_i \otimes \Pm^c_i$. Oftentimes, finding such a canonical absolute pose $\T_i$ from a single local patch involves computing \textit{local reference frames}~\cite{salti2014shot}, that are known to be unreliable~\cite{petrelli2011repeatability}. We instead base our idea on the premise that a good local (patch-wise) pose estimation leads to a good global rigid alignment of two fragments. First, by decoupling the pose component from the structure information, we devise a data driven predictor network capable of regressing the pose for arbitrary patches and showing good generalization properties. Fig.~\ref{fig:pipelineRelNet} depicts our architectural design. In a following part, we tackle the problem of relative pose labeling without the need for a canonical frame computation. %\cSlo{We should stress that we also obtain an improved keypoint descriptor in this way.}\cTolga{This is just the method part, should we still market it?} \cSlo{Probably not at this place, but maybe later}

\paragraph{Generalized pose prediction}
\label{sec:prediction}
A naive way to achieve tolerance to 3D-structure is to train the network for pose prediction conditioned on a database of input patches and leave the invariance up to the network~\cite{zeng20163dmatch,deng2018ppfnet}. Unfortunately, networks trained in this manner either demand a very large collection of unique local patches or simply lack generalization. To alleviate this drawback, we opt to eliminate the structural components by training an invariant-equivariant network pair and using the intermediary latent space arithmetic.
We characterize an equivariant function $\Psi$ as~\cite{worrall2018}:
\begin{align}
    \Psi(\Pm) = \Psi(\T \otimes \Pm^c) = g(\T) \Psi(\Pm^c)\label{eq:equiv}
\end{align}
where $g(\cdot)$ is a function dependent only upon the pose. When $g(\T)=\mathbf{I}$, $\Psi$ is said to be $\T$-\textit{invariant} and for the scope of our application, for any input $\Pm$ leads to the outcome of the canonical one $\Psi(\Pm)\gets \Psi(\Pm^c)$. Note that~\cref{eq:equiv} is more general than Cohen's definition~\cite{cohen2016group} as the group element $\T$ is not restricted to act linearly. Within the body of this paper the term \textit{equivariant} will loosely refer to such \textit{quasi-equivariance} or \textit{co-variance}.
When $g(\T)\neq\mathbf{I}$, we further assume that the action of $\T$ can be approximated by some additive linear operation:
\begin{align}
    g(\T) \Psi(\Pm^c) \approx h(\T) + \Psi(\Pm^c).\label{eq:approx}
\end{align}
$h(\T)$ being a probably highly non-linear function of $\T$. By plugging~\cref{eq:approx} into~\cref{eq:equiv}, we arrive at:
\begin{equation}
\Psi(\Pm)-\Psi(\Pm^c)\approx h(\T)
\end{equation}
that is, the difference in the latent space can approximate the pose up to a non-linearity, $h$. We approximate the inverse of $h$ by a four-layer MLP network $h^{-1}(\cdot)\triangleq\rho(\cdot)$ and propose to regress the motion (rotational) terms:
\begin{equation}
     \rho(\fb) \approx \Rot \,|\, \mathbf{t}
\end{equation}
where $\fb=\Psi(\Pm)-\Psi(\Pm^c)$. Note that $\fb$ solely explains the motion and hence, can generalize to any local patch structure, leading to a powerful pose predictor under our mild assumptions.

The manifolds formed by deep networks are found sufficiently close to a Euclidean flatness. This rather flat nature has already motivated prominent works such as GANs~\cite{goodfellow2014generative} to use simple latent space arithmetic to modify faces, objects etc. Our assumption in \cref{eq:approx} follows a similar premise. Semantically speaking, by \textit{subtracting out} the structure specific information from point cloud features, we end up with descriptors that are pose/motion-focused.

\paragraph{Relative pose estimation}
\label{sec:pose}
Note that $\rho(\cdot)$ can be directly used to regress the absolute pose to a canonical frame. Yet, due to the aforementioned difficulties of defining a unique local reference frame, it is not advised~\cite{petrelli2011repeatability}. 
Since our scenario considers a pair of scenes, we can safely estimate a \textit{relative pose} rather than the absolute, ousting the prerequisite for a nicely estimated LRF. This also helps us to easily forge the labels needed for training. Thus, we model $\rho(\cdot)$ by a relative pose predictor, \textit{RelativeNet}, as shown in~\cref{fig:pipelineRelNet}.

We further make the observation that, correspondent local structures of two scenes $(i,j)$ that are well-registered under a rigid transformation $\T_{ij}$ also align well with $\T_{ij}$. As a result, the relative pose between local patches could be easily obtained by calculating the relative pose between the fragments and vice versa. We will use these ideas in the following section~\cref{sec:network} to design our networks, and in~\cref{sec:training} explain how to train them. 

\subsection{Network Design}
\label{sec:network}
To realize our generalized relative pose prediction, we need to implement three key components: the invariant network $\Psi(\Pm^c)$ where $g(\T)=\mathbf{I}$, the network $\Psi(\Pm)$ that varies as a function of the input and the MLP $\rho(\cdot)$.
The recent PPF-FoldNet~\cite{Deng_2018_ECCV} auto-encoder is luckily very suitable to model $\Psi(\Pm^c)$, as it is unsupervised, works on point patches and achieves true invariance thanks to the point pair features (PPF) fully marginalizing the motion terms. Interestingly, keeping the network architecture identical as PPF-FoldNet, if we were to substitute the PPF part with the 3D points themselves ($\Pm$), the intermediate feature would be dependent upon both structure and pose information. We coin this version as \textit{PC-FoldNet} and use it as our equivariant network $\Psi(\Pm)=g(\T)\Psi(\Pm^c)$.
We rely on using PPF-FoldNet and PC-FoldNet to learn rotation-invariant and -variant features respectively. They share the same architecture while take in a different encoding of local patches, as shown in~\cref{fig:foldnet}. Taking the difference of the encoder outputs of the two networks, i.e. the latent features of PPF- and PC-FoldNet respectively, results in new features which specialize almost exclusively on the pose (motion) information. Those features are subsequently fed into the generalized pose predictor RelativeNet to recover the rigid relative transformation. 
The overall architecture of our complete relative pose prediction is illustrated in~\cref{fig:pipelineRelNet}.
\insertimageC{1}{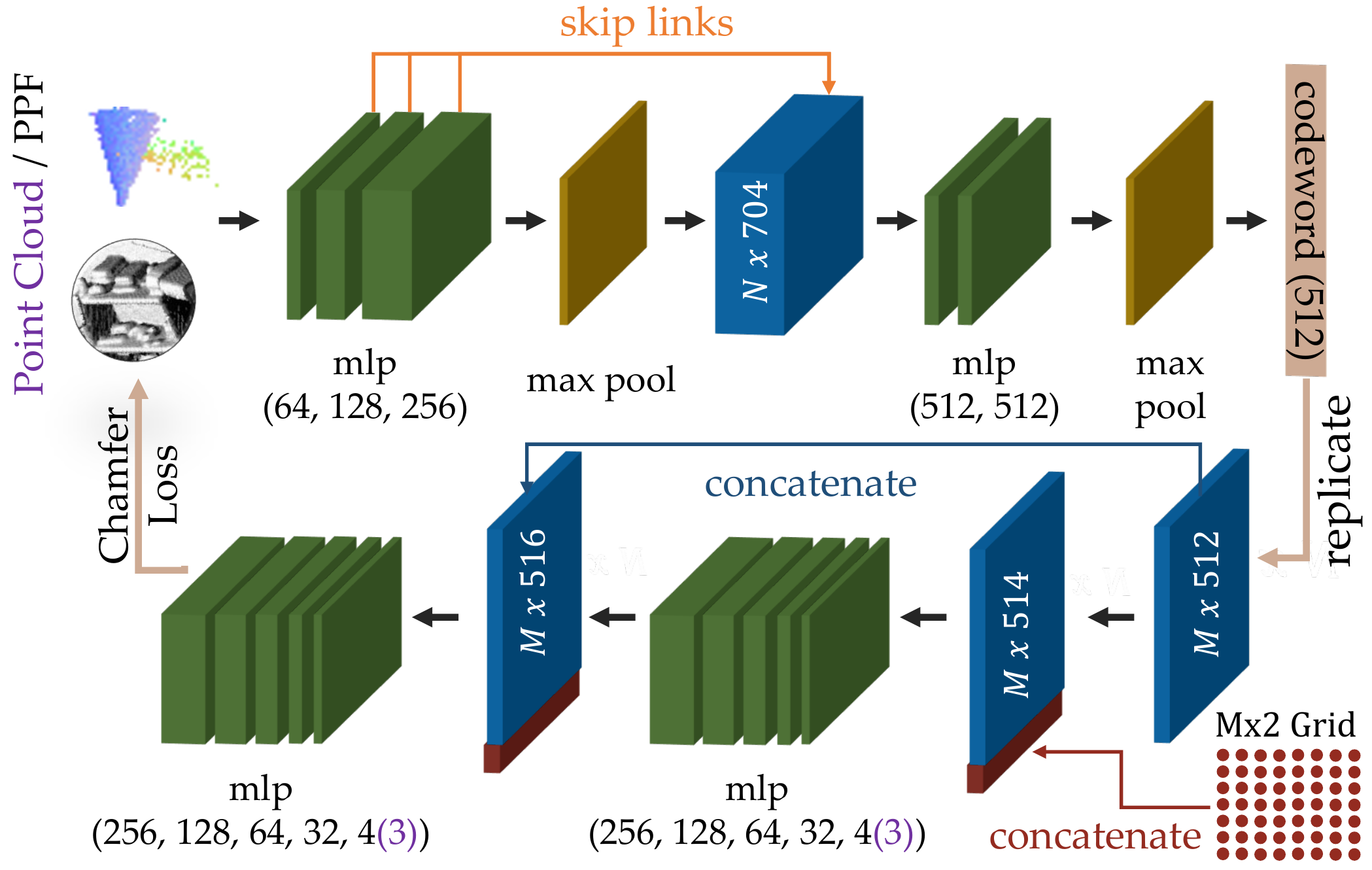}{The architecture of PC/PPF-FoldNet. Depending on the input source, the number of last layers of unfolding module is 3 for point clouds and 4 for point pair features, respectively.\vspace{-2mm}}{fig:foldnet}{t!}

%Since we aim at predicting relative poses, we get a pair of subtracted intermediate features from the pair patches. They are concatenated as the input to a RotNet to predict their relative poses. For simplicity, RotNet is implemented as a 4-layer MLP.

\subsection{Multi-Task Training Scheme}
\label{sec:training}
We train our networks with multiple cues, supervised and unsupervised. In particular, our loss function $L$ is  composed of three parts:
\begin{equation}
    L = L_{rec} +\lambda_1 L_{pose} + \lambda_2L_{feat}
\end{equation}
$L_{rec}$, $L_{pose}$ and $L_{feat}$ are the reconstruction, pose prediction and feature consistency losses, respectively. For the sake of clarity, we omit the function arguments. 
%During the training, multiple training signals are combined to guide the network to find the optimized solution for our pose regression task. 
\paragraph{Reconstruction loss} $L_{rec}$ reflects the reconstruction fidelity of PC/PPF-FoldNet. To enable the encoders of PPF/PC-FoldNet to generate good features for pose regression, as well as for finding robust local correspondences, similar to the steps in PPF-FoldNet\cite{Deng_2018_ECCV}, use the \textit{Chamfer Distance} as the metric to train the both of the auto-encoders in an unsupervised manner:
\begin{align}
    &L_{rec} =\frac{1}{2}\Big( d_{cham}(\Pm, \hat{\Pm}) + d_{cham}(\F_{ppf}, \hat{\F}_{ppf})\Big)\\
&d_{cham}(\X,\hat{\X}) = \\
&\text{max}\Bigg\{ \frac{1}{|\X|}\sum\limits_{\x \in \X} \min_{\hat{\x}\in \hat{\x}} \| \x-\hat{\x} \|_2, \,\, \frac{1}{|\hat{\X}|}\sum\limits_{\hat{\x} \in \hat{\X}} \min_{\x\in\X} \| \x-\hat{\x} \|_2 \Bigg\}.\nonumber
\end{align}
$\,\hat{}\,$ operator denotes the reconstructed (estimated) set and $\F_{ppf}$ the PPFs of the points computed identically as~\cite{Deng_2018_ECCV}.
\paragraph{Pose prediction loss}
A correspondence of two local patches are centralized and normalized before being sent into PC/PPF-FoldNets. This cancels the translational part $\tb \in \R^3$. The main task of our pose prediction loss is then to enable our RelativeNet to predict the relative rotation $\mathbf{R}_{12} \in SO(3)$ between given patches $(1,2)$. Hence, a natural choice for $L_{pose}$ describes the discrepancy between the predicted and the ground truth rotations:
\begin{equation}
\label{eq:q}
L_{pose} = \lVert \q - \q^* \rVert_2
\end{equation}
Note that we choose to parameterize the spatial rotations by quaternions $\q\in \mathbb{H}_{1}$, the Hamiltonian 4-tuples~\cite{busam2017camera,birdal2018bayesian} due to: 1) decreased the number of parameters to regress, 2) lightweight projection operator - vector-normalization. 

Translation $\tb^*$, conditioned on the hypothesized pair $(\p_1, \p_2)$ and the predicted rotation $\q^*$ can be computed by:
\begin{equation}
\label{eq:t}
\tb^* = \p_1 - \mathbf{R}^{*}\p_2
\end{equation}
where $\mathbf{R}^{*}$ corresponds to the matrix representation of $\q^*$. Such an L2 error is easier to train with negligible loss compared to the geodesic metric.
\paragraph{Feature consistency loss}
Unlike~\cite{Deng_2018_ECCV}, our RelativeNet requires pairs of local patches for training. Thus, we can additionally make use of pair information as an extra \textit{weak supervision} signal to further facilitate the training of our PPF-FoldNet. We hypothesize that such guidance would improve the quality of intermediate latent features that were previously trained in a fully unsupervised fashion. In specific, correspondent features subject to noise, missing data or clutter would generate a high reconstruction loss causing the local features to be different even for the same local patches. 
This new information helps us to guarantee that the features extracted from identical patches live as close as possible in the embedded space, which is extremely beneficial since we establish local correspondences by searching their nearest neighbor in the feature space. The \textit{feature consistency loss} $L_{feat}$ reads:
\begin{equation}
    L_{feat} = \sum_{(\p_i, \q_i) \in \mathbf{\Gamma}}\lVert \fb_{\p_i} - \fb_{\q_i} \rVert_2
\end{equation}
$\mathbf{\Gamma}$ represents the set of correspondent local patches and $\fb_{\p}$ is the feature extracted at $\p$ by the PPF-FoldNet, $\fb_{\p}\in \F_{ppf}$.

% \subsection{Rigid Transformation Parameterization}

% Quaternions vs Rodrigues rotation vector

\subsection{Hypotheses Selection}
\label{sec:hypo}
The final stage of our algorithm involves selecting the best hypotheses among many, produced per each sample point.
The full 6DoF pose is parameterized by the predicted 3DoF orientation (\cref{eq:q}) and the translation (\cref{eq:t}) conditioned on matching points (3DoF). For our approach, having a set of correspondences is equivalent to having a pre-generated set of transformation hypotheses since each keypoint is associated an LRF. Note that this is contrary to the standard RANSAC approaches where $m=3$-correspondences parameterize the pose, and establishing $N$ correspondences can lead to ${N}\choose{m}$ hypotheses to be verified. 
Our small number of hypotheses, already linear in the number of correspondences, makes it possible to exhaustively evaluate the putative matching pairs for verification. We further refine the estimate by recomputing the transformation using all the surviving inliers. The hypothesis with the highest score would be kept as the final decision.

~\cref{fig:prediction} shows that both translational and rotational components of our hypothesis set are tighter and have smaller deviation from the true pose as opposed to the standard RANSAC hypotheses.

\insertimageC{1}{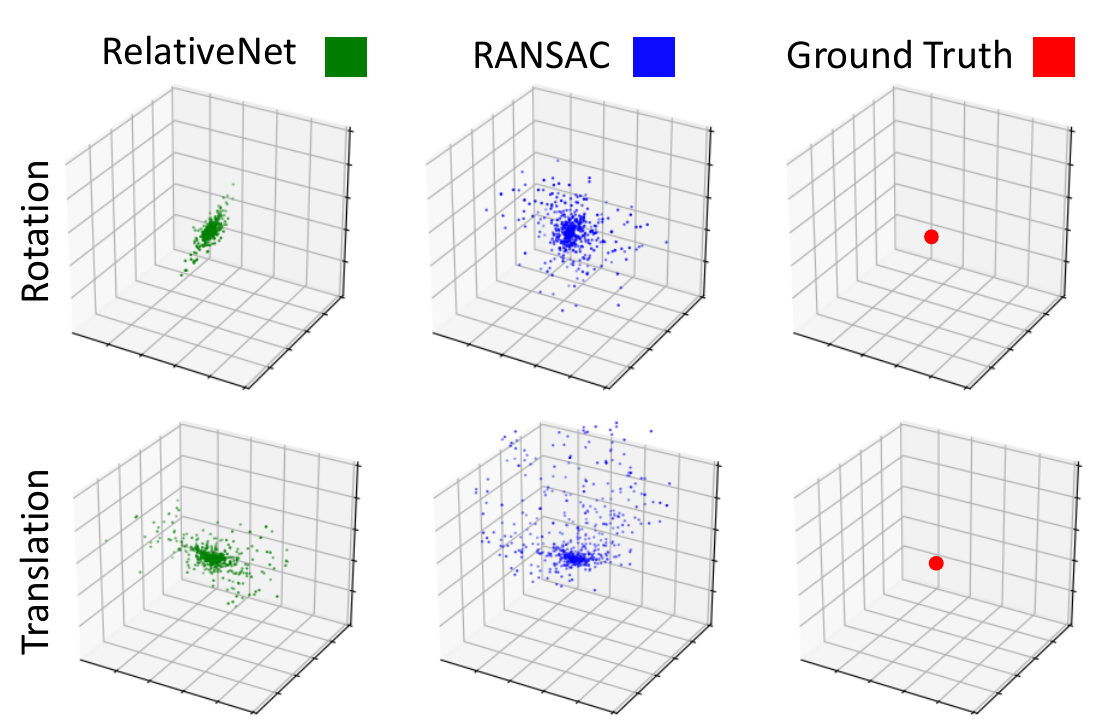}{Comparison between the hypotheses generated by our Direct Prediction and RANSAC pipeline. The first row shows the rotational component as 3D Rodrigues vectors, and the second row shows the translational component. Hypotheses generated by our RelativeNet are more centralized around the ground truth.\vspace{-3mm}}{fig:prediction}{t!}
\begin{table*}[t!]
  \centering
  \caption{Results on 3DMatch benchmark for fragment matching recall~\cite{zeng20163dmatch,Deng_2018_ECCV}.}
  \resizebox{\linewidth}{!}{\small
    \begin{tabular}{lccccccccc}
    \toprule\small
          & Kitchen  & Home 1  & Home 2  & Hotel 1  & Hotel 2  & Hotel 3  & Study  & MIT Lab  & Average  \\
\midrule
\midrule
3DMatch~\cite{zeng20163dmatch}  & 0.5751 & 0.7372 & 0.7067 & 0.5708 & 0.4423 & 0.6296 & 0.5616 & 0.5455 & 0.5961 \\
      CGF~\cite{Khoury_2017_ICCV}  & 0.4605 & 0.6154 & 0.5625 & 0.4469 & 0.3846 & 0.5926 & 0.4075 & 0.3506 & 0.4776 \\
      PPFNet~\cite{deng2018ppfnet}  &  \textbf{0.8972}  & 0.5577 & 0.5913 & 0.5796 & 0.5769 & 0.6111 & 0.5342 & 0.6364 & 0.6231 \\
      FoldingNet~\cite{Yang_2018_CVPR}  & 0.5949 & 0.7179 & 0.6058 & 0.6549 & 0.4231 & 0.6111 & 0.7123 & 0.5844 & 0.613 \\
      PPF-FoldNet~\cite{Deng_2018_ECCV}  & 0.7352 & 0.7564 & 0.625 & 0.6593 & 0.6058 & 0.8889 & 0.5753 & 0.5974 & 0.6804 \\
      \midrule
      Ours & 0.7964 &  \textbf{0.8077} &  \textbf{0.6971} &  \textbf{0.7257} &  \textbf{0.6731} &   \textbf{0.9444} &  \textbf{0.6986} &  \textbf{0.6234} &  \textbf{0.7458} \\
      \bottomrule
    \end{tabular}\vspace{-3mm}
  \label{tab:3dmatchbenchmark}%
  }
\end{table*}%

\section{Experiments}
\label{sec:experiments}
We train our method using the training split of the de-facto 3DMatch benchmark dataset~\cite{zeng20163dmatch}, containing lots of real local patch pairs with different structure and pose, captured by Kinect cameras. We then conduct evaluations on its own test set and on the challenging synthetic Redwood Benchmark~\cite{Choi_2015_CVPR}. We assess our performance against the state of the art data-driven algorithms as well as the prosperous handcrafted methods of the RANSAC-family on the tasks of feature matching and geometric registration.
\vspace{-3mm}\paragraph{Implementation details}
We represent a local patch by randomly collecting $2$K points around a reference one within $30$cm vicinity.
To provide relative pose supervision, we associate each patch a pose fetched from the ground truth relative transformations.
Local correspondences are established by finding the mutually closest neighbors in the feature space. Our implementation is based on PyTorch~\cite{pytorch}, a widely used deep learning framework. 
%\subsection{3D Match Benchmark~\cite{zeng20163dmatch}}
\subsection{Evaluations on 3D Match Benchmark~\cite{zeng20163dmatch}}
%3D Match Benchmark is now the de-facto standard for evaluating deeply learned local features. It also provides a training set, where lots of local patch pairs with different structures and poses are generated for network training. 
\paragraph{How good are our local descriptors?}
We begin by putting our local features at test for fragment matching task, which reflects how many good correspondence sets could be found by the specific features. A fragment pair is said to match if a true correspondence ratio of $5\%$ and above is achieved. See~\cite{Deng_2018_ECCV,deng2018ppfnet} for details. In~\cref{tab:3dmatchbenchmark} we report the recall of various data driven descriptors, 3DMatch~\cite{zeng20163dmatch}, CGF~\cite{Khoury_2017_ICCV}, PPFNet~\cite{deng2018ppfnet}, FoldingNet~\cite{Yang_2018_CVPR}, PPF-FoldNet~\cite{Deng_2018_ECCV}, as well as ours. It is remarkable to see that our network outperforms the supervised PPFNet~\cite{deng2018ppfnet} by $\sim 12\%$ and the unsupervised PPF-FoldNet~\cite{Deng_2018_ECCV} by $\sim 6\%$. Note that, we are architecturally identical to PPF-FoldNet and hence the improvement is enabled primarily by the multi-task training signals, interacting towards a better minimum and decoupling of the shape and pose within the architecture. Thanks to the double-siamese structure of our network, we can provide both rotation-invariant features like~\cite{Deng_2018_ECCV}, or upright ones, similar to~\cite{deng2018ppfnet}.

\begin{table*}[htbp]
  \centering
  \caption{Geometric registration performance comparison. The first part lists the performances of some state-of-the-art deeply learned local features combined with RANSAC. The second part shows the performances of our features combined with RANSAC and its variants. The third part shows the results of our features combined with our pose prediction module directly. Not only our learned features are more powerful, but also our pose prediction module demonstrates superiority over RANSAC family.}
  \resizebox{\linewidth}{!}{%
    \begin{tabular}{cc|ccccccccc|c}
    \toprule
          & \multicolumn{1}{c}{} &       &  Kitchen   &  Home 1   &  Home 2   &  Hotel 1   &  Hotel 2   &  Hotel 3   &  Study   &  MIT Lab   &  Average   \\
    \midrule
    \midrule
    \multicolumn{1}{c|}{\multirow{6}[2]{*}{\shortstack{Different\\Feautures \\+\\RANSAC}}} & \multirow{2}[1]{*}{\shortstack{3DMatch\\~\cite{zeng20163dmatch}}} & Rec.  & 0.8530 & 0.7830 & 0.6101 & 0.7857 & 0.5897 & 0.5769 & 0.6325 & 0.5111 & 0.6678 \\
    \multicolumn{1}{c|}{} &       & Prec. & 0.7213 & 0.3517 & 0.2861 & 0.7186 & 0.4144 & 0.2459 & 0.2691 & 0.2000 & 0.4009 \\
    \multicolumn{1}{c|}{} & \multirow{2}[0]{*}{\shortstack{CGF\\~\cite{Khoury_2017_ICCV}}} & Rec.  & 0.7171 & 0.6887 & 0.4591 & 0.5495 & 0.4872 & 0.6538 & 0.4786 & 0.4222 & 0.5570 \\
    \multicolumn{1}{c|}{} &       & Prec. & 0.5430 & 0.1830 & 0.1241 & 0.3759 & 0.1538 & 0.1574 & 0.1605 & 0.1033 & 0.2251 \\
    \multicolumn{1}{c|}{} & \multirow{2}[1]{*}{\shortstack{PPFNet\\~\cite{deng2018ppfnet}}} & Rec.  & \bf{0.9020} & 0.5849 & 0.5723 & 0.7473 & 0.6795 & 0.8846 & 0.6752 & 0.6222 & 0.7085 \\
    \multicolumn{1}{c|}{} &       & Prec. & 0.6553 & 0.1546 & 0.1572 & 0.4159 & 0.2181 & 0.2018 & 0.1627 & 0.1267 & 0.2615 \\
    \midrule
    \multicolumn{1}{c|}{\multirow{8}[2]{*}{\shortstack{Our\\Features \\+\\ RANSAC \\variants}}} & \multirow{2}[1]{*}{\shortstack{USAC\\~\cite{raguram2013usac}}} & Rec.  & 0.8820 & 0.7642 & 0.6101 & 0.7527 & 0.6538 & 0.8077 & 0.6709 & 0.5778 & 0.7149 \\
    \multicolumn{1}{c|}{} &       & Prec. & 0.5083 & 0.1397 & 0.1362 & 0.2972 & 0.1536 & 0.1329 & 0.1530 & 0.1053 & 0.2033 \\
    \multicolumn{1}{c|}{} & \multirow{2}[0]{*}{\shortstack{SPRT\\~\cite{chum2008optimal}}} & Rec.  & 0.8797 & 0.7453 & 0.6101 & 0.7253 & 0.6538 & 0.8462 & 0.6624 & 0.4444 & 0.6959 \\
    \multicolumn{1}{c|}{} &       & Prec. & 0.5170 & 0.1341 & 0.1374 & 0.3158 & 0.1599 & 0.1384 & 0.1593 & 0.0881 & 0.2062 \\
    \multicolumn{1}{c|}{} & \multirow{2}[0]{*}{\shortstack{LR\\~\cite{korman2018latent}}} & Rec.  & 0.8753 & 0.7925 & 0.6038 & 0.7198 & \bf{0.7051} & 0.7692 & 0.6667 & 0.5556 & 0.7110 \\
    \multicolumn{1}{c|}{} &       & Prec. & 0.5019 & 0.1348 & 0.1294 & 0.2854 & 0.1549 & 0.1190 & 0.1465 & 0.1012 & 0.1967 \\
    \multicolumn{1}{c|}{} & \multicolumn{1}{c|}{\multirow{2}[1]{*}{\shortstack{RAN\\SAC}}} & Rec.  & 0.8530 & 0.7642 & 0.6038 & 0.7033 & 0.6667 & 0.7692 & 0.6496 & 0.5111 & 0.6901 \\
    \multicolumn{1}{c|}{} &       & Prec. & 0.5527 & 0.1614 & 0.1479 & 0.3647 & 0.1825 & 0.1587 & 0.1658 & 0.1139 & 0.2309 \\
    \midrule
    \multicolumn{2}{c|}{\multirow{2}[2]{*}{\shortstack{Our Features +\\ Pose Prediction}}} & Rec.  & 0.8998 & \bf{0.8302} & \bf{0.6352} & \bf{0.8242} & 0.6923 & \bf{0.9231} & \bf{0.7650} & \bf{0.6444} & \bf{0.7768} \\
    \multicolumn{2}{c|}{} & Prec. & 0.5437 & 0.1778 & 0.1807 & 0.4011 & 0.2061 & 0.2087 & 0.1843 & 0.1465 & 0.2561 \\
    \bottomrule
    \end{tabular}}\vspace{-3mm}
  \label{tab:registration}%
\end{table*}%

\paragraph{How useful are our features in geometric registration?} To further demonstrate the superiority of our learned local features, we evaluate them for the task of local geometric registration (L-GM). In a typical L-GM pipeline, local features are first extracted and then a set of local correspondences are established by some form of a search in the latent space. Out of these putative matches, a subsequent RANSAC iteratively selects a subset of minimally 3 correspondences in order to estimate a rigid pose. The best relative rigid transformation between the fragment pair is then the one with the highest inlier score. For the sake of fairness among all the methods and to have a controlled setting where the result depends only on the differences in descriptors, we use the simple RANSAC framework~\cite{raguram2013usac} across all methods to find the best matches. 

The first part of~\cref{tab:registration} shows how well different local features could aid RANSAC to register fragments on the 3DMatch Benchmark. Recall and precision are computed the same way as in 3DMatch~\cite{zeng20163dmatch}. For this evaluation, recall is a more important measure, because the precision can be improved by employing better hypothesis pruning schemes filtering out the bad matches without harming recall~\cite{korman2018latent,Khoury_2017_ICCV}. The registration result shows that our method is on par with or better than the best performer PPFNet~\cite{deng2018ppfnet} on average recall, while using a much more light-weighted training pipeline. Interestingly, our recall on this task drops when compared to the one of the fragment matching. This means that for certain fragment pairs, even though the inlier ratio is above $5\%$, RANSAC fails to do the work. Thus, one is motivated to seek better ways to recover the rigid transformation from 3D correspondences. 

\vspace{-3mm}\paragraph{How accurate is our direct 6D prediction?} 
We now evaluate the contributions of RelativeNet in fixing the aforementioned breaking cases of RANSAC. Thanks to our architecture, we are able to endow each correspondence with a pose information. Normally, each of these correspondences are expected to be good. However, in practice this is not the case. Hence, we devise a linear search to find the best of those, as explained in~\cref{sec:hypo}. In~\cref{tab:registration} (bottom), we report our L-GM results as an outcome of this verification, on the same 3DMatch Benchmark. As we can see, with the same set of correspondences, our method could yield a much higher recall, reaching up to $77.68\%$, around $8\%$ higher than what is achievable by RANSAC. This is $7\%$ higher than PPFNet. Also, this number is around $3\%$ higher than the recall in fragment matching, which means that not only pairs with good correspondences are registered, but also some challenging pairs with even less than $5\%$ inlier ratio are successfully registered, pushing the potential of matched correspondences to the limit. 

It is noteworthy to point out that the iterative scheme of RANSAC requires finding at least 3 correct correspondences to estimate $\mathbf{T}$, whereas it is sufficient for us to rely on a single correct match. Moreover due to downsampling~\cite{birdal2017sampling}, poses computed directly from 3-points are crude, whereas patch-wise pose predictions of our network are less prone to the accuracy of exact keypoint location. 
% We report local geometric result of our method with direct 6D pose estimation. As shown in Tab.~\ref{tab:3dregistration}, combined with the powerful descriptors learned by our framework, our direct Pose Prediction could boost the recall up to 77.68\%, around 8\% higher than using RANSAC. Also this result outperforms all the other competitors by large margin. 
\vspace{-2.5mm}\paragraph{Comparisons against the RANSAC-family} 
To further demonstrate the power of RelativeNet, we compare it with some of the state-of-the-art variants of RANSAC, namely USAC~\cite{raguram2013usac}, SPRT~\cite{chum2008optimal} and Latent RANSAC (LR)~\cite{korman2018latent}. Those methods are proved to be both faster and more powerful than the vanilla version~\cite{raguram2013usac, korman2018latent}. 
% We base our implementation on the USAC~\cite{raguram2013usac} framework, integrating all methods in a modular fashion. 
%We also introduce another baseline using USAC with SPRT\cite{chum2008optimal}.
%SPRT is also an extension for RANSAC, providing an alternative for verifying hypotheses. For briefness, we tag the USAC with SPRT extension as SPRT. Another recent work is called LatentRANSAC (abbreviated as LR), which utilizes Random Hashing Grids to detect collision to decrease the number of hypotheses to be verified, thus accelerating the RANSAC pipeline. 

All the methods are given the same set of putative matching points found by our rotation-invariant features. The results depicted in~\cref{tab:registration} shows that even a simple hypothesis prunning combined with our data driven RelativeNet can surpass an entire set of hand-crafted methods, achieving approximately $6.19\%$ higher reacall than the best obtained by USAC and $2.61\%$ better than the highest precision obtained by standard RANSAC. In this regard, our method takes a dominant advantage on 3D pairwise geometric registration.
\insertimageStar{1}{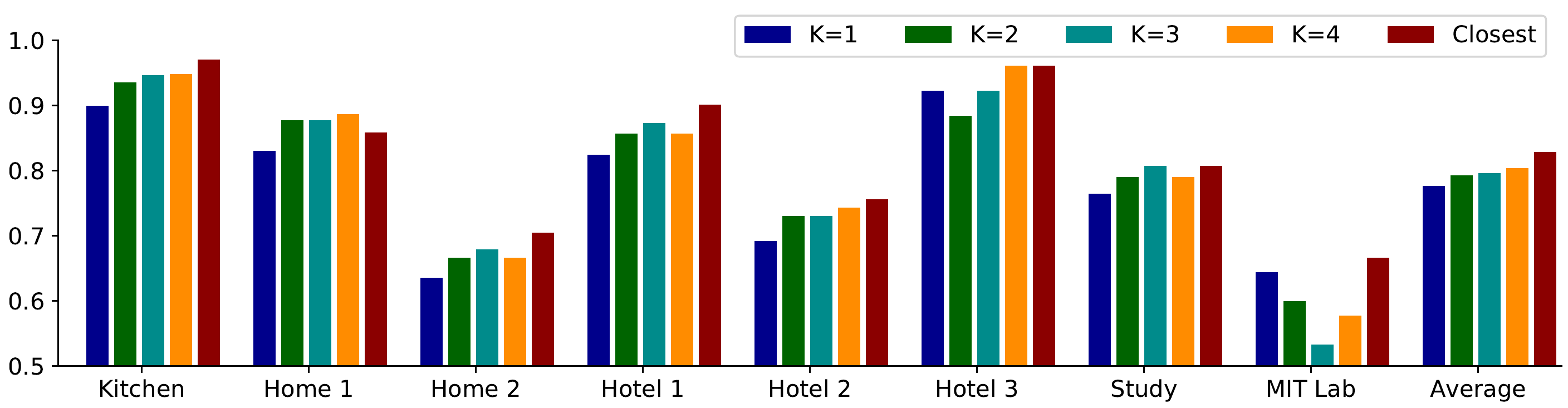}{The impact of using different methods to find correspondences. As the number of mutual correspondences kept, $K$, increases, more hypotheses are verified leading to a trade-off between recall and computation time.\vspace{-3mm}}{fig:correspondence}{t!}
\vspace{-9.5mm}\paragraph{Running times}
Speed is  another important factor regarding any pairwise registration algorithm and it is of interest to see how our work compares to the state of the art in this aspect. We implement our hypotheses verification part based on USAC to make the comparison fair with other USAC-based implementations.

The average time needed for registering a fragment pair is recorded in~\cref{tab:runtime}, feature extraction time excluded. All timings are done on a Intel(R) Core(TM) i7-4820K CPU @ 3.70GHz with a single thread. Note that, our method is much faster than the fastest RANSAC-variant \textit{Latent-RANSAC}~\cite{korman2018latent}. The average time for generating all hypotheses for a fragment pair by RelativeNet is about 0.013s, and the subsequent verification costs 0.016s, making up around 0.03s in total.  An important reason why we can terminate so quickly is that the number of hypotheses generated and verified is much smaller compared to the RANSAC methods. While LR is capable of reducing this amount significantly, the number of surviving hypotheses to be verified is still much more than ours. 

\begin{table}[hbtp]
  \centering
  \caption{The average runtime for registering one fragment pair and the number of hypotheses generated and verified.}
   \resizebox{\columnwidth}{!}{
    \begin{tabular}[\columnwidth]{ccccc}
    \toprule
    & USAC~\cite{raguram2013usac} & SPRT~\cite{chum2008optimal} & LR~\cite{korman2018latent} & Ours\\
    \midrule
    Time(s) & 0.886 & 2.661 & 0.591 & 0.013 + 0.016\\
    \# Hypos & 30220 & 672223 & 2568 (46198) & 335\\
    \bottomrule
    \end{tabular}
    }
  \label{tab:runtime}\vspace{-4mm}
\end{table}%

% Quote:
%  here in 3dmatch dataset, the average number of hypothesis to be verified is around ~300, the average time to verify is around 0.02. But as the number of hypothesis grow, the number of calculations grows exponentially. So if we have ~10000 hypothesis, the time would be around $0.02 \times (33)^2  = 20$s. Then it's a typical time we would need in a redwood dataset case, or close to this level.

% \subsection{Ablation Study}
% \paragraph{How does each component in the loss function impact feature quality?}
% TODO:

% \paragraph{How does architecture design help pose prediction?}
% TODO: 

\insertimageC{1}{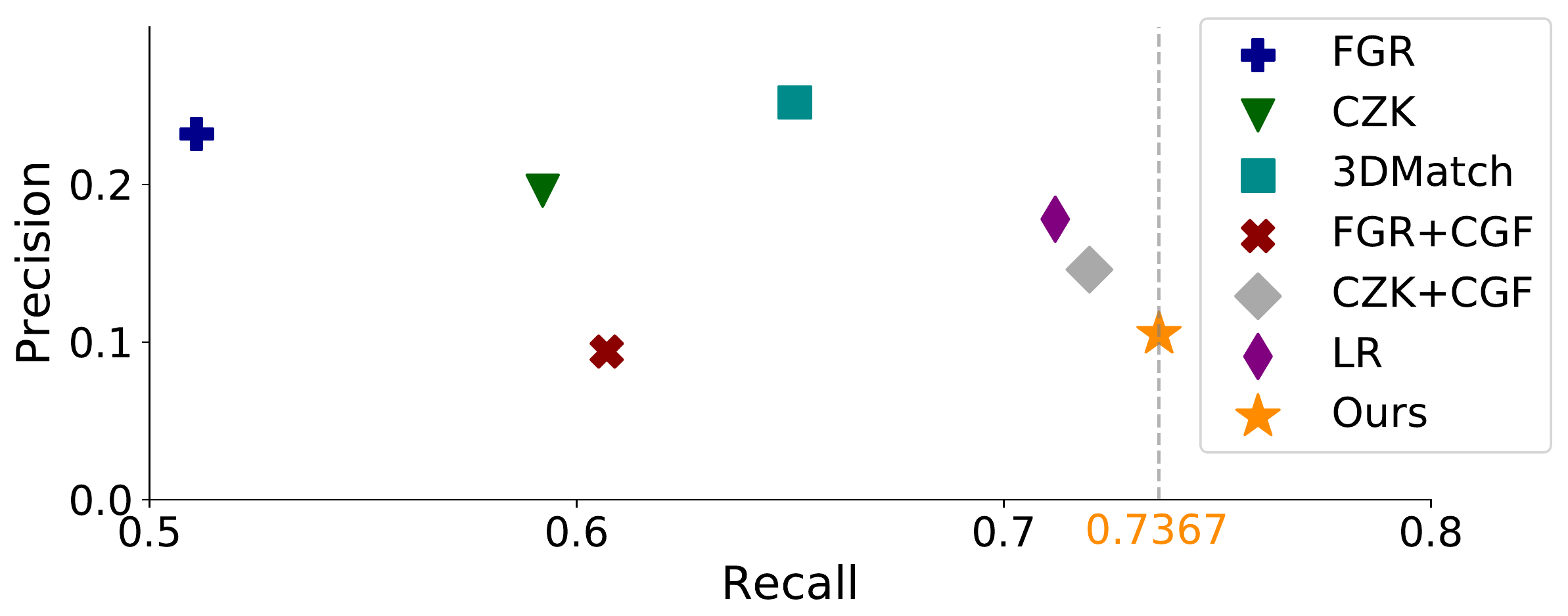}{Geometric registration performance of various methods on Redwood Benchmark~\cite{Choi_2015_CVPR}.\vspace{-4mm}}{fig:redwood}{t!}

\paragraph{Effect of correspondence estimation on the registration}
We put 5 different ways to constructing putative matching pair sets under an ablation study. Strategies include: (1) keeping different number of mutual closest neighboring patches $k=1\dots4$, each dubbed as {$K=k$} and (2) keeping a nearest neighbor for all the local patches from both fragments as a match pair, dubbed \textit{Closest}. These strategies are applied on the same set of local features to estimate initial correspondences for further registration. The results of each method on different scenes and their average are plotted in~\cref{fig:correspondence}. As $k$ increases and the criteria for accepting a neighbor to be a pair relaxes, we observe an overall trend of increasing registration recall on different sequences. Not surprisingly, this trend is most obvious in the \textit{Average} column. This is of course not sufficient to conclude that relaxation helps correspondences.
The second important observation is that the number of established correspondences also increases as this condition relaxes. 
The average amount of putative matches found by \textit{Closest} is around 3664, much larger than $K=1$'s 334, approximately $10$ times more, meaning that a subsequent verification would need more time to process them. Hence, we arrive at the conclusion that if recall/accuracy is the main concern, more putative matches should be kept. If, conversely, speed is an issue, \textit{Mutual-1} could achieve a rather satisfying result quicker.

%\subsection{Generalization to Unseen Data}
\vspace{-2.5mm}\paragraph{Generalization to unseen domains} 
To show that our algorithm could generalize well to other datasets, we evaluate its performance on the well-known and challenging global registration benchmark provided by Choi \etal, the Redwood Benchmark~\cite{Choi_2015_CVPR}. This dataset contains four different synthetic scenes with sequence of fragments. Our network is not fine-tuned with any synthetic data, instead, the weights trained with real data from 3DMatch dataset is used directly. We follow the evaluation settings as Choi~\etal for an easy and fair comparison, and report the registration results in~\cref{fig:redwood}. This precision and recall plot also depcits results achieved by some recent methods including FGR~\cite{zhou2016fast}, CZK~\cite{Choi_2015_CVPR}, 3DMatch~\cite{zeng20163dmatch}, CGF+FGR~\cite{Khoury_2017_ICCV}, CGF+CZK~\cite{Khoury_2017_ICCV}, and Latent-Ransac~\cite{korman2018latent}. Among them, 3DMatch and CGF are data-driven. 3DMatch was trained with real data on the same data source as ours, while CGF trained with synthetic data. Note that our method shows $\sim8.5\%$ higher recall against 3DMatch. Although we are not using any synthetic data for finetuning, we still achieve a better recall of $2.4\%$ w.r.t. CGF and its combination with CZK. In general, our method outperforms all the other state-of-the-art methods on Redwood Benchmark~\cite{Choi_2015_CVPR}, which validates the generalizability and good performance of our method simultaneously. Note that while in general, the maximal precision is low across all the methods, it is not hard to improve it when the recall is high. To show that recall is the primary measure, we ran a global optimization~\cite{Choi_2015_CVPR} on our initial results, bringing precision up to $91\%$ without big loss of recall - still at $73\%$.

% TODO:

% \subsection{Generalization to Synthetic Dataset}

\vspace{-1.5mm}\section{Conclusion}\vspace{-1.0mm}
\label{sec:conclusion}
We proposed a unified end-to-end framework for both local feature extraction and pose prediction. Comprehensive experiments on 3DMatch benchmark demonstrate that a multi-task training scheme could inject more power into the learned features, hence improve the quality of the correspondence set for further registration. Geometric registration using the pose predictions by our RelativeNet given the putative matched pairs is also shown to be both more robust and much faster than various state-of-the-art RANSAC methods. We also studied how different methods of establishing local correspondences would affect the registration performance. The outstanding performance on the challenging synthetic Redwood benchmark strongly validates that our method is not only robust, but also generalizes well to unseen datasets. In the future, we also plan to introduce a data driven hypotheses verification approach.
%Our method opens a new possibility of using deep networks to solve local geometric registration in a robust and efficient style.

{\small
%\bibliographystyle{ieee}
%\bibliography{egbib}

}

% UNCOMMENT THIS TO COMPILE WITH APPENDIX
\setcounter{section}{0}
\renewcommand\thesection{\Alph{section}}
\newcommand{\suppsection}{\subsection}
\clearpage
\section{Appendix}

\begin{figure}[b!]
\centering
\subfloat[Equivariant Feature Matching]{
\includegraphics[width=0.7\linewidth, clip=true]{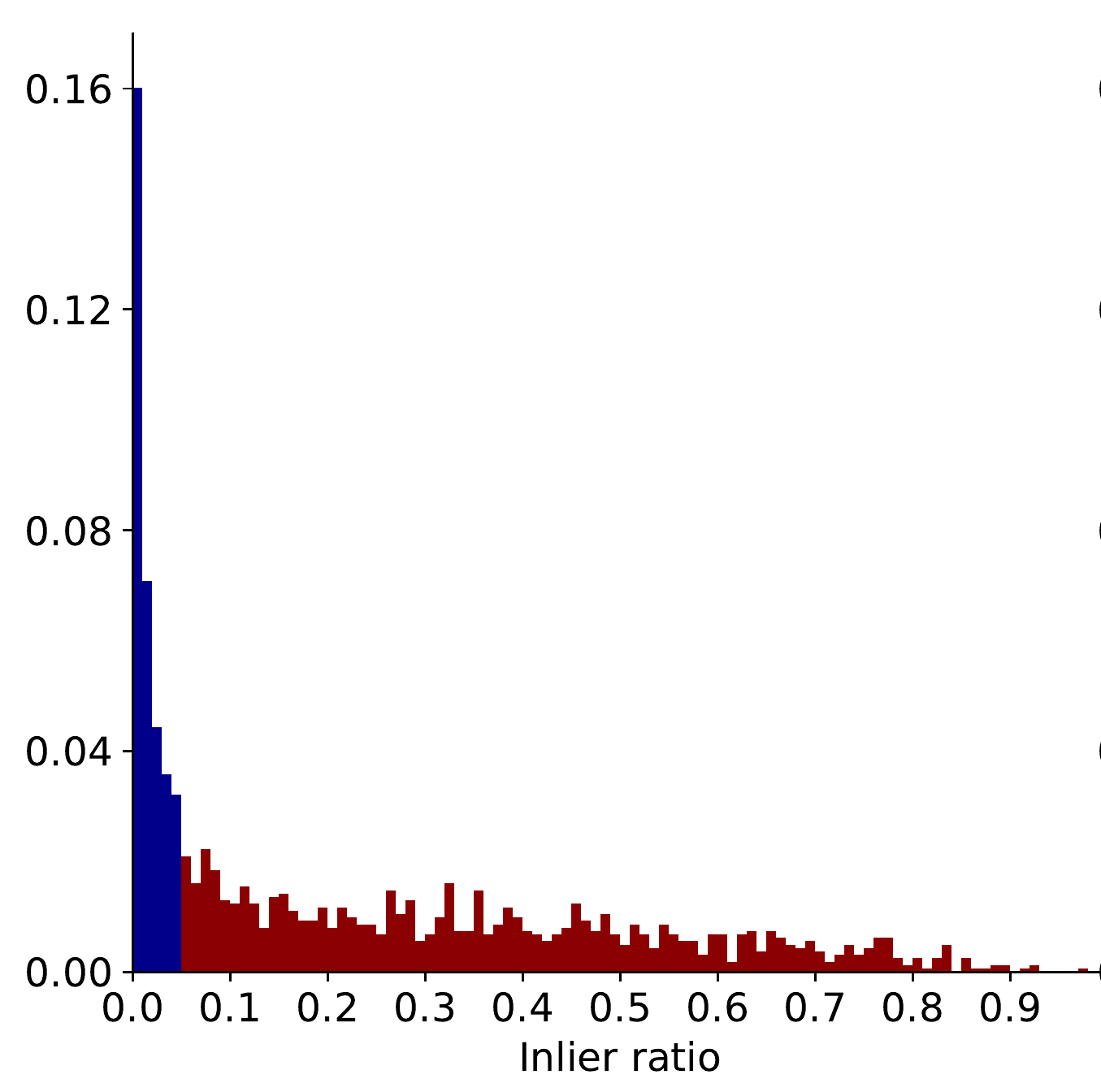}
\label{subfig:equiv}}\\
\subfloat[Invariant Feature Matching]{
\includegraphics[width=0.7\linewidth, clip=true]{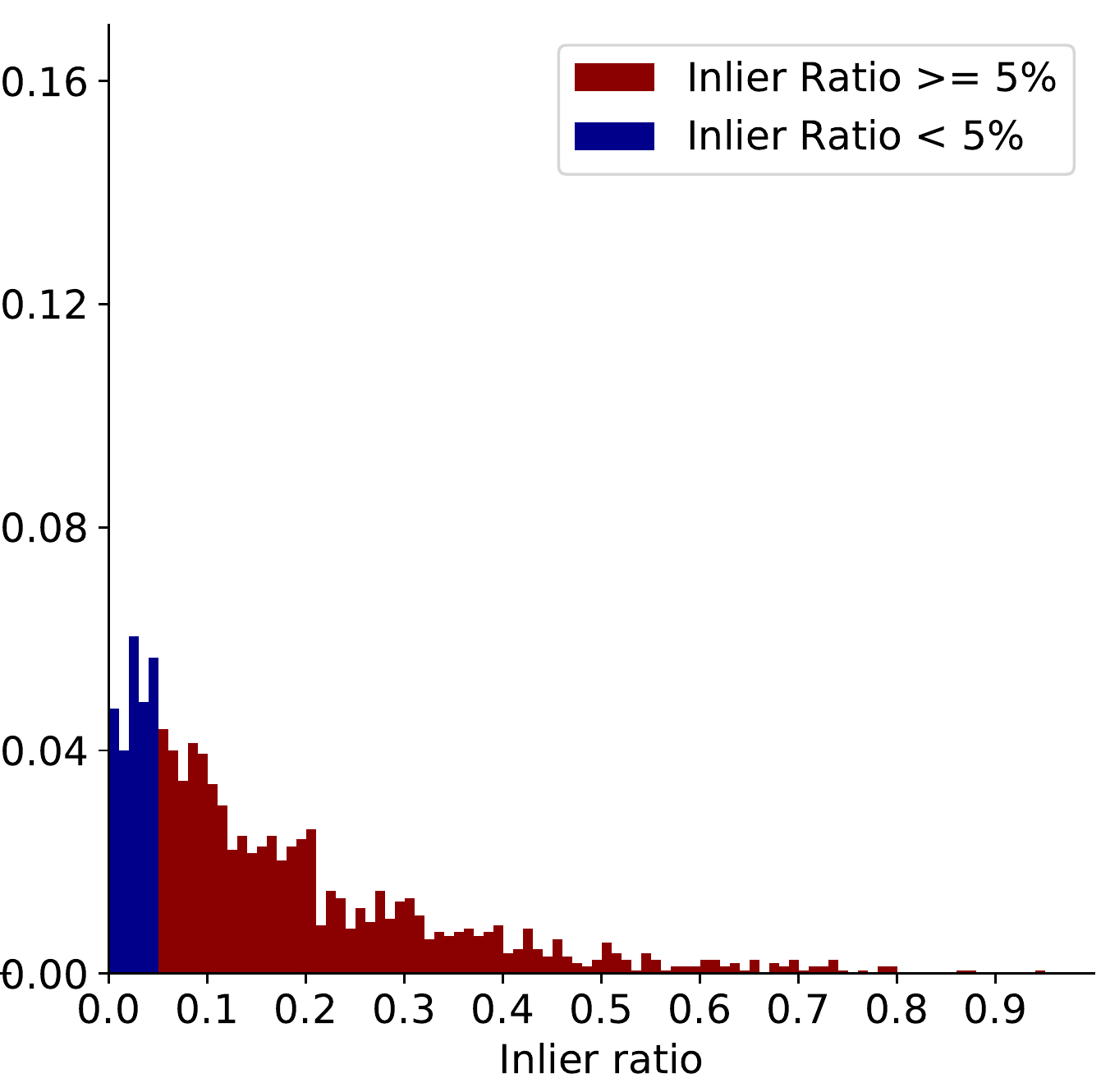}
\label{subfig:invar}}
\caption{Inlier ratio distribution of fragment pair matching result using different local features from our framework. \textbf{(a)} Matching results using equivariant features extracted by PC-FoldNet. \textbf{(b)} Matching results using invariant features extracted by PPF-FoldNet. Blue part stands for the portion of fragment pairs with correspondence inlier ratio smaller than 5\%. Matching results by invariant features demonstrate a better quality for further registration procedure.} 
%\vspace{-0.5em}
\label{fig:feature}
\end{figure}

\insertimageStar{1}{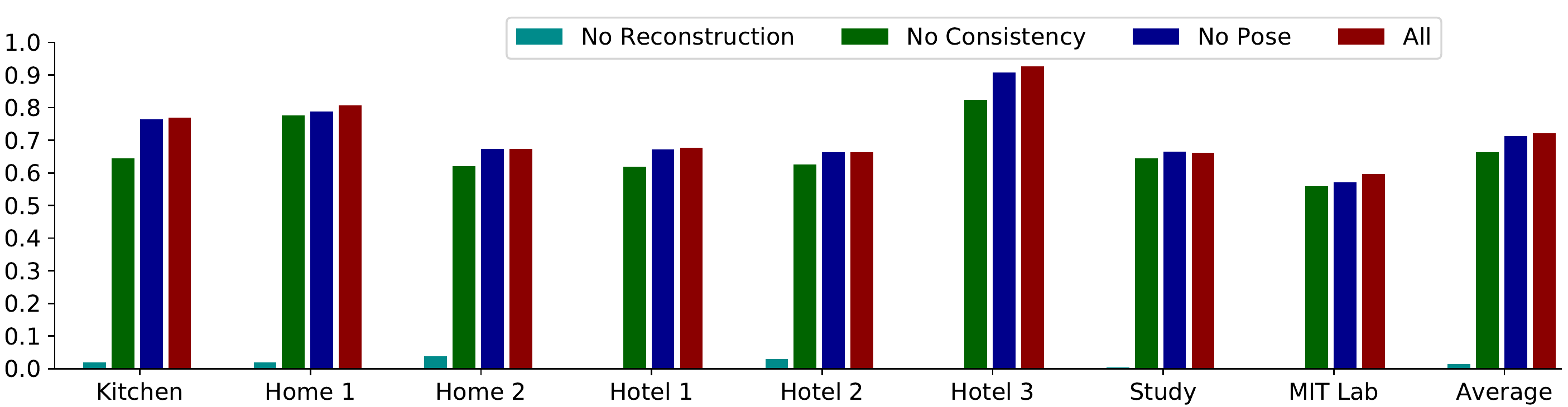}{Influences of different supervision signals. {Reconstruction} is the most essential loss for our network to generate local features for matching tasks. Without it the descriptive-ness is lost. When all losses are combined, the network learns to extract the most powerful features and achieves the best performance.}{fig:loss}{t!}

\subsection{Ablation Study}
\paragraph{Does multi-task training scheme help to boost the feature quality?} In order to find out how multi-task training affects the quality of the learned intermediate features, we trained several networks with combinations of different supervision signals. For the sake of controlled experimentation, all networks are made to have the identical architecture. They are trained with the same data for 10 epochs. Hence, the only variable remains to be the objective function used for each group. 

In total, there are four networks to be compared. The first one is trained with all the available supervision signals, i.e. reconstruction loss, feature consistency loss and pose prediction loss. Regarding the other three groups, each of the networks is trained with one of the three signals excluded. For simplicity, those groups are tagged as \textit{All}, \textit{No Reconstruction},  \textit{No Consistency} and \textit{No Pose} respectively. The fragment matching results using features from different networks are shown in~\cref{fig:loss}.

As shown in \cref{fig:loss}, with all the training signals on, the learned features are the most robust and outperform all the others which lack at least one piece of information and thus suffer a performance drop. When no reconstruction loss is applied, the learned features almost always fail at matching. It is therefore the most critical loss to minimize. The absence of pose prediction loss has the least negative influence. Yet, it is necessary for RelativeNet to learn to predict the relative pose for given patch pairs. Without this the later stages of the pipeline such as hypotheses generation and verification cannot continue. These results validate that our multi-task training scheme takes full advantage of all the available information to drive the performance of learned local features to a higher level. 

\paragraph{Matching of invariant vs pose-variant features} Our method extracts two kinds of local features using two different network components. The ones extracted by PPF-FoldNet are fully rotation-invariant, while local features of PC-FoldNet change as the pose of local patches vary. Experimentation contained in the paper used local features from PPF-FoldNet only to establish correspondences thanks to its superior property of invariance. Here, we use invariant and equivariant features to match fragment pairs separately, and compare their matching performance. This is important in validating our choice that invariant features are more suitable for nearest neighbor queries.

~\cref{fig:feature} exhibits the distribution of correspondence inlier ratio for the matched fragment pairs by using different local features. Matching results of equivariant features shows a huge amount of fragment pairs having correspondences with only a small fraction of inliers (less than ~5\%). Invariant features though, manage to provide many fragment pairs with a set of correspondences with over 10\% true matches. It proves that invariant features are better at finding good correspondence set for further registration stage. All in all, rotation-invariant features extracted by PPF-FoldNet is more suitable for finding putative local matches. Note that this was also verified by ~\cite{Deng_2018_ECCV}.  
\begin{table}[h!]
  \centering
  \caption{Average \# of correspondences obtained by different methods of assignments. $K=k$ refers to retaining $k$-mutual neighbors.}
    \resizebox{\columnwidth}{!}{\small\begin{tabular}{lccccc}
    \toprule
          & $K=1$ & $K=2$ & $K=3$ & $K=4$ & Closest \\
    \midrule
    \# Matches & 335   & 1099  & 1834  & 2609  & 3664 \\
    \bottomrule
    \end{tabular}%
    }
  \label{tab:correspondences}%
\end{table}
\paragraph{More details for correspondence estimation methods} 
In the main paper, we found out that a more relaxed condition for keeping neighbors lead to a better subsequent registration. However, this performance gain comes at a cost and hence introduces a trade-off.~\cref{tab:correspondences} tabulates the average number of putative matches found by different methods. As we can see, the size of correspondence set increases rapidly as we relax the standard and keep more neighbors. In return, this means more computation time in the following registration stage.

\begin{figure*}[t!]
\centering
\subfloat[Rotation Part]{
\includegraphics[width=0.5\linewidth, clip=true]{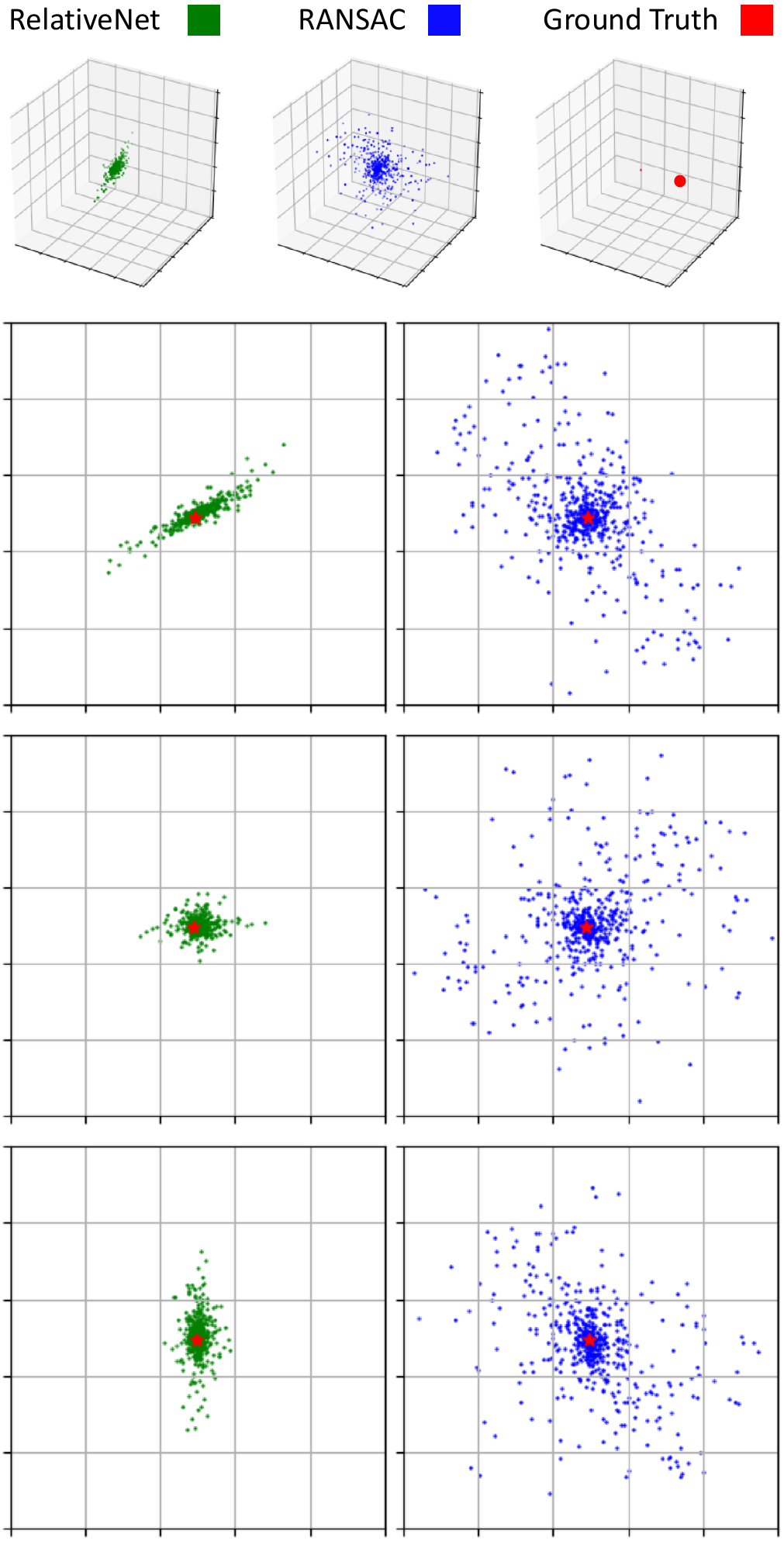}
\label{subfig:dist_pc}}
\subfloat[Translation Part]{
\includegraphics[width=0.5\linewidth, clip=true]{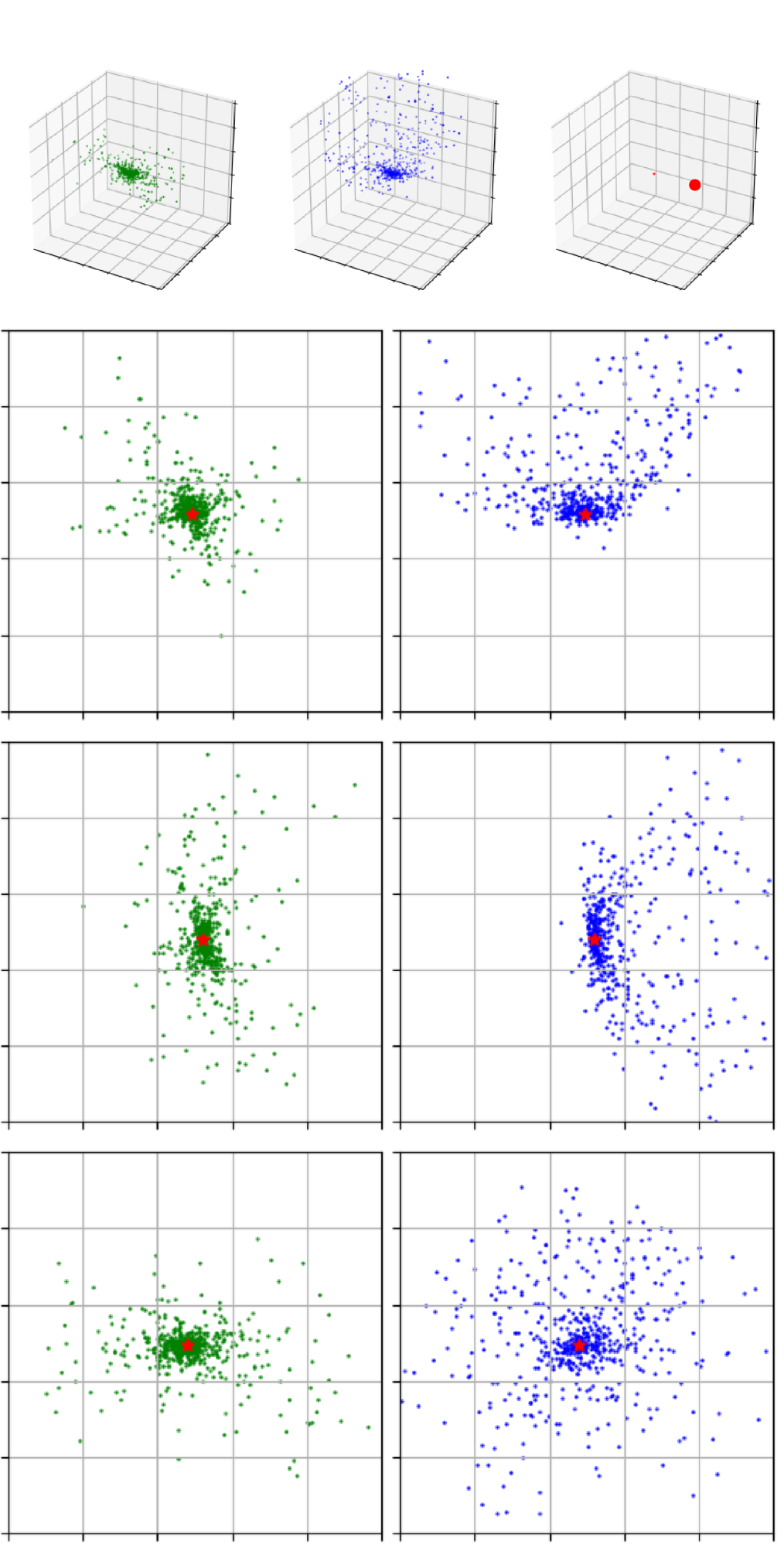}
\label{subfig:dist_ppf}}
\caption{Hypotheses distribution comparison between ones generated by RANSAC using randomly selected  subset of correspondences and ones predicted by our RelativeNet. Rotation and translation parts are shown separately. The first row plots the distributions in 3D space and the following three rows are correspondent 2D projections from three different orthogonal view directions. } 
%\vspace{-0.5em}
\label{fig:distribution}
\end{figure*}

\insertimageStar{1}{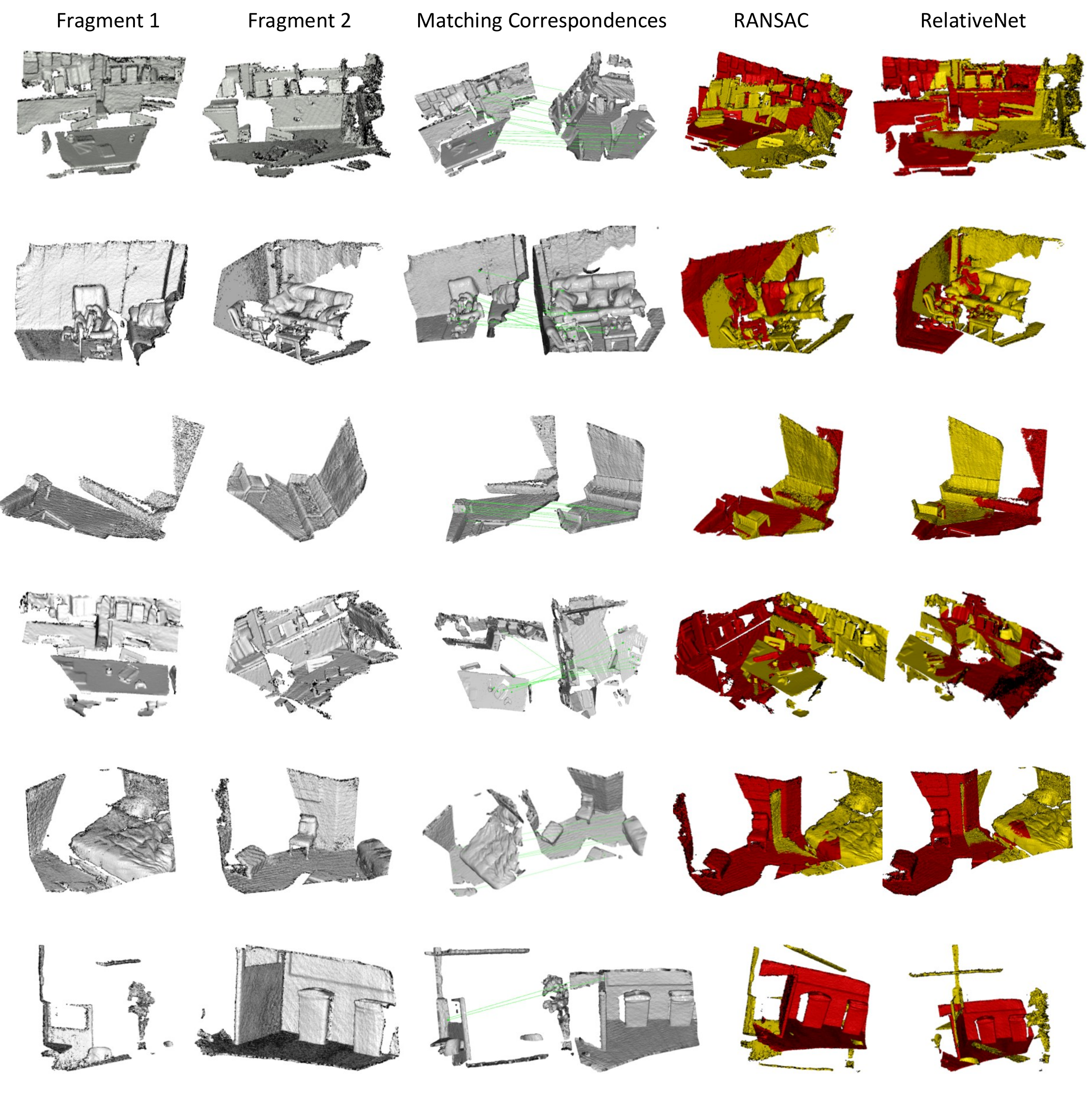}{Some challenging fragment pairs with only a small number of correct correspondences. RANSAC fails to estimate the correct relative poses between them while our network is able to produce successful registration results. Especially, for the fragment pair in the last row, only two correct local correspondences are found, which doesn't satisfy the minimum number of inliers required by RANSAC, but still correctly handled by our method.}{fig:match_comp}{t!}
\subsection{Quantitative Results}
\paragraph{Distribution of hypotheses}
~\cref{fig:distribution} shows the distribution of poses predicted by RelativeNet and poses determined by running RANSAC on the randomly selected subsets of corresponding points. Each hypothesis is composed of a rotational and translational part. The former is represented as a Rodrigues vector to keep it in $\mathbb{R}^3$. It is obvious that hypotheses predicted by RelativeNet are centered more around the ground truth pose, both in rotation and translation. It also reveals the reason why the hypotheses of our network could facilitate an easier and faster registration procedure. 

\paragraph{Qualitative comparison against RANSAC}
~\cref{fig:match_comp} shows some challenging cases where only a small number of correct correspondences are established. In these examples, RANSAC fails to recover the pose information from the small set of inliers hidden in a big set of mismatches. However, a registration procedure with the aid of RelativeNet could succeed with a correct result. The qualitative comparison demonstrates that our method is robust at registering fragment pairs even in extreme cases where insufficient inliers are presented. 

\paragraph{Multi-scan registration}
Finally, we apply our method in registering multiple scans to a common reference frame. To do that, we first align pairwise scans and obtain the most likely relative pose per pair. These poses are then fed into a global registration pipeline~\cite{choi2015robust}. Note that while this method can use a global iterative closest point alignment~\cite{besl1992method} in the final stage, we deliberately omit this step to emphasize the quality of our pairwise estimates. Hence, the outcome is a rough, but nevertheless an acceptable alignment on which we can optionally apply the global-ICP refining the points and scans. The results are shown in~\cref{fig:redkitchen} on the \textit{Red Kitchen} sequence of the 7-scenes~\cite{shotton2013scene} as well as in~\cref{fig:sun3dhotel} on the Sun3D Hotel sequence~\cite{xiao2013sun3d}, a part of 3DMatch benchmark~\cite{zeng20163dmatch}.

\insertimageStar{1}{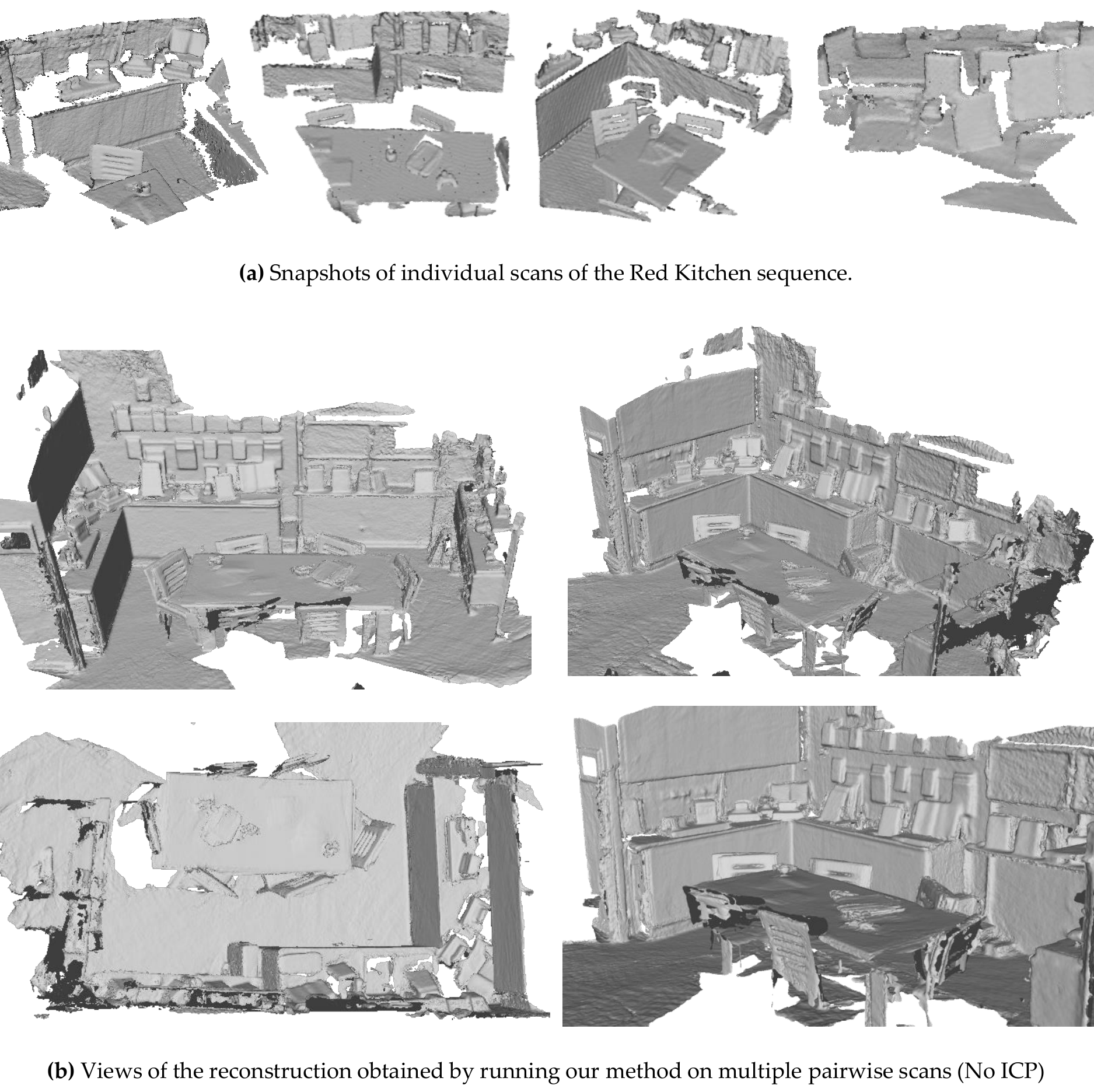}{Reconstruction by 3D alignment on the entire Red Kitchen sequence of the 7scenes dataset~\cite{shotton2013scene}. We first compute the pairwise estimates by our method and feed them into the pipeline of~\cite{choi2015robust} for obtaining the poses in a globally coherent frame. Note that this dataset is a real one, acquired by a Kinect scanner. We make no assumptions on the order of acquisition.}{fig:redkitchen}{ht}

\insertimageStar{1}{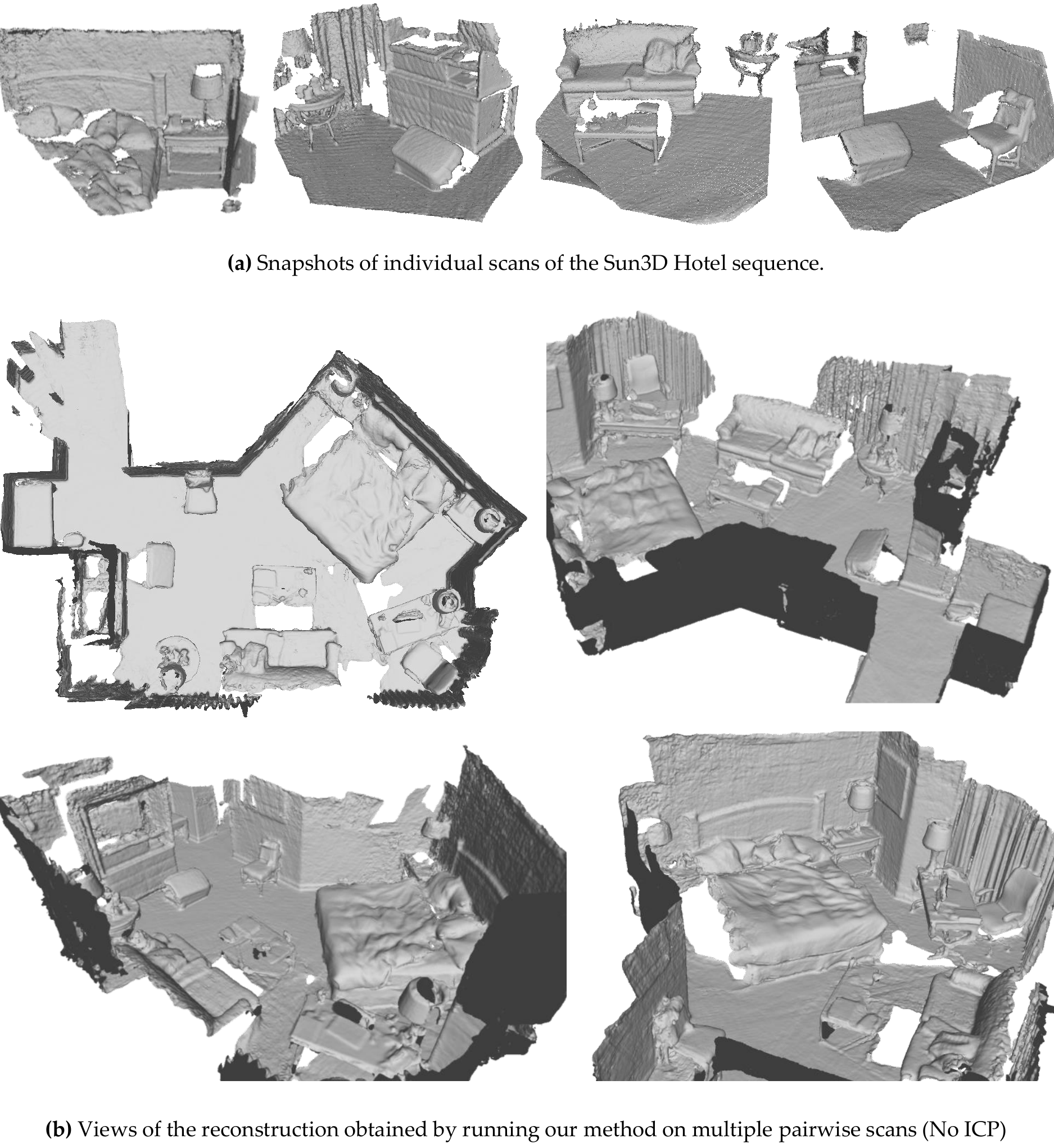}{Reconstruction by 3D alignment on the entire Sun3D Hotel sequence. The reconstruction procedure is identical to the one of~\cref{fig:redkitchen}.}{fig:sun3dhotel}{ht}
% 
% \onecolumn{
% \bibliographystyle{ieee}
% \bibliography{egbib}

\begin{thebibliography}{10}\itemsep=-1pt

\bibitem{aiger20084}
D.~Aiger, N.~J. Mitra, and D.~Cohen-Or.
\newblock 4-points congruent sets for robust pairwise surface registration.
\newblock In {\em ACM Transactions on Graphics (TOG)}, volume~27, page~85. ACM,
  2008.

\bibitem{besl1992method}
P.~J. Besl and N.~D. McKay.
\newblock Method for registration of 3-d shapes.
\newblock In {\em Robotics-DL tentative}, pages 586--606. International Society
  for Optics and Photonics, 1992.

\bibitem{birdal2016online}
T.~Birdal, E.~Bala, T.~Eren, and S.~Ilic.
\newblock Online inspection of 3d parts via a locally overlapping camera
  network.
\newblock In {\em 2016 IEEE Winter Conference on Applications of Computer
  Vision (WACV)}, pages 1--10. IEEE, 2016.

\bibitem{birdal3dv2015}
T.~Birdal and S.~Ilic.
\newblock Point pair features based object detection and pose estimation
  revisited.
\newblock In {\em 3D Vision}, pages 527--535. IEEE, 2015.

\bibitem{birdal2017cad}
T.~Birdal and S.~Ilic.
\newblock Cad priors for accurate and flexible instance reconstruction.
\newblock In {\em 2017 IEEE International Conference on Computer Vision
  (ICCV)}, pages 133--142, Oct 2017.

\bibitem{birdal2017point}
T.~Birdal and S.~Ilic.
\newblock A point sampling algorithm for 3d matching of irregular geometries.
\newblock In {\em 2017 IEEE/RSJ International Conference on Intelligent Robots
  and Systems (IROS)}, pages 6871--6878. IEEE, 2017.

\bibitem{birdal2017sampling}
T.~Birdal and S.~Ilic.
\newblock A point sampling algorithm for 3d matching of irregular geometries.
\newblock In {\em International Conference on Intelligent Robots and Systems
  (IROS 2017)}. IEEE, 2017.

\bibitem{birdal2018bayesian}
T.~Birdal, U.~Simsekli, M.~O. Eken, and S.~Ilic.
\newblock Bayesian pose graph optimization via bingham distributions and
  tempered geodesic mcmc.
\newblock In {\em Advances in Neural Information Processing Systems}, pages
  306--317, 2018.

\bibitem{bueno20184}
M.~Bueno, F.~Bosch{\'e}, H.~Gonz{\'a}lez-Jorge, J.~Mart{\'\i}nez-S{\'a}nchez,
  and P.~Arias.
\newblock 4-plane congruent sets for automatic registration of as-is 3d point
  clouds with 3d bim models.
\newblock {\em Automation in Construction}, 89:120--134, 2018.

\bibitem{busam2017camera}
B.~Busam, T.~Birdal, and N.~Navab.
\newblock Camera pose filtering with local regression geodesics on the
  riemannian manifold of dual quaternions.
\newblock In {\em Proceedings of the IEEE International Conference on Computer
  Vision}, pages 2436--2445, 2017.

\bibitem{bustos2018guaranteed}
{\'A}.~P. Bustos and T.-J. Chin.
\newblock Guaranteed outlier removal for point cloud registration with
  correspondences.
\newblock {\em IEEE transactions on pattern analysis and machine intelligence},
  40(12):2868--2882, 2018.

\bibitem{choi1997performance}
S.~Choi, T.~Kim, and W.~Yu.
\newblock Performance evaluation of ransac family.
\newblock {\em Journal of Computer Vision}, 24(3):271--300, 1997.

\bibitem{Choi_2015_CVPR}
S.~Choi, Q.-Y. Zhou, and V.~Koltun.
\newblock Robust reconstruction of indoor scenes.
\newblock In {\em IEEE Conference on Computer Vision and Pattern Recognition
  (CVPR)}, 2015.

\bibitem{choi2015robust}
S.~Choi, Q.-Y. Zhou, and V.~Koltun.
\newblock Robust reconstruction of indoor scenes.
\newblock In {\em Proceedings of the IEEE Conference on Computer Vision and
  Pattern Recognition}, 2015.

\bibitem{chum2005matching}
O.~Chum and J.~Matas.
\newblock Matching with prosac-progressive sample consensus.
\newblock In {\em Computer Vision and Pattern Recognition, 2005. CVPR 2005.
  IEEE Computer Society Conference on}, volume~1, pages 220--226. IEEE, 2005.

\bibitem{chum2008optimal}
O.~Chum and J.~Matas.
\newblock Optimal randomized ransac.
\newblock {\em IEEE Transactions on Pattern Analysis and Machine Intelligence},
  30(8):1472--1482, 2008.

\bibitem{chum2003locally}
O.~Chum, J.~Matas, and J.~Kittler.
\newblock Locally optimized ransac.
\newblock In {\em Joint Pattern Recognition Symposium}, pages 236--243.
  Springer, 2003.

\bibitem{cohen2016group}
T.~Cohen and M.~Welling.
\newblock Group equivariant convolutional networks.
\newblock In {\em International conference on machine learning}, pages
  2990--2999, 2016.

\bibitem{Deng_2018_ECCV}
H.~Deng, T.~Birdal, and S.~Ilic.
\newblock Ppf-foldnet: Unsupervised learning of rotation invariant 3d local
  descriptors.
\newblock In {\em The European Conference on Computer Vision (ECCV)}, September
  2018.

\bibitem{deng2018ppfnet}
H.~Deng, T.~Birdal, and S.~Ilic.
\newblock Ppfnet: Global context aware local features for robust 3d point
  matching.
\newblock {\em Computer Vision and Pattern Recognition (CVPR). IEEE}, 1, 2018.

\bibitem{drost2010model}
B.~Drost, M.~Ulrich, N.~Navab, and S.~Ilic.
\newblock Model globally, match locally: Efficient and robust 3d object
  recognition.
\newblock In {\em Computer Vision and Pattern Recognition (CVPR), 2010 IEEE
  Conference on}, pages 998--1005. Ieee, 2010.

\bibitem{Eckart_2018_ECCV}
B.~Eckart, K.~Kim, and J.~Kautz.
\newblock Hgmr: Hierarchical gaussian mixtures for adaptive 3d registration.
\newblock In {\em The European Conference on Computer Vision (ECCV)}, September
  2018.

\bibitem{Fischler1981}
M.~A. Fischler and R.~C. Bolles.
\newblock Random sample consensus: A paradigm for model fitting with
  applications to image analysis and automated cartography.
\newblock {\em Commun. ACM}, 1981.

\bibitem{Gojcic2019}
Z.~Gojcic, C.~Zhou, J.~D. Wegner, and W.~J. D.
\newblock The perfect match: 3d point cloud matching with smoothed densities.
\newblock In {\em IEEE Conf. on Computer Vision and Pattern Recognition
  (CVPR)}, June 2019.

\bibitem{goodfellow2014generative}
I.~Goodfellow, J.~Pouget-Abadie, M.~Mirza, B.~Xu, D.~Warde-Farley, S.~Ozair,
  A.~Courville, and Y.~Bengio.
\newblock Generative adversarial nets.
\newblock In {\em Advances in neural information processing systems}, pages
  2672--2680, 2014.

\bibitem{guo2014performance}
Y.~Guo, M.~Bennamoun, F.~Sohel, M.~Lu, J.~Wan, and J.~Zhang.
\newblock Performance evaluation of 3d local feature descriptors.
\newblock In {\em Asian Conference on Computer Vision}, pages 178--194.
  Springer, 2014.

\bibitem{hinterstoisser2016going}
S.~Hinterstoisser, V.~Lepetit, N.~Rajkumar, and K.~Konolige.
\newblock Going further with point pair features.
\newblock In {\em European Conference on Computer Vision}, pages 834--848.
  Springer, 2016.

\bibitem{bop}
T.~Hoda{\v{n}}, F.~Michel, E.~Brachmann, W.~Kehl, A.~G. Buch, D.~Kraft,
  B.~Drost, J.~Vidal, S.~Ihrke, X.~Zabulis, C.~Sahin, F.~Manhardt, F.~Tombari,
  T.-K. Kim, J.~Matas, and C.~Rother.
\newblock Bop: Benchmark for 6d object pose estimation.
\newblock In V.~Ferrari, M.~Hebert, C.~Sminchisescu, and Y.~Weiss, editors,
  {\em Computer Vision -- ECCV 2018}, pages 19--35, 2018.

\bibitem{jian20183dfeat}
Z.~Jian~Yew and G.~Hee~Lee.
\newblock 3dfeat-net: Weakly supervised local 3d features for point cloud
  registration.
\newblock In {\em Proceedings of the European Conference on Computer Vision
  (ECCV)}, pages 607--623, 2018.

\bibitem{johnson1999using}
A.~E. Johnson and M.~Hebert.
\newblock Using spin images for efficient object recognition in cluttered 3d
  scenes.
\newblock {\em IEEE Transactions on pattern analysis and machine intelligence},
  21(5):433--449, 1999.

\bibitem{Lawin_2018_CVPR}
F.~Järemo~Lawin, M.~Danelljan, F.~Shahbaz~Khan, P.-E. Forssén, and
  M.~Felsberg.
\newblock Density adaptive point set registration.
\newblock In {\em The IEEE Conference on Computer Vision and Pattern
  Recognition (CVPR)}, June 2018.

\bibitem{kabsch1976solution}
W.~Kabsch.
\newblock A solution for the best rotation to relate two sets of vectors.
\newblock {\em Acta Crystallographica Section A: Crystal Physics, Diffraction,
  Theoretical and General Crystallography}, 32(5):922--923, 1976.

\bibitem{Khoury_2017_ICCV}
M.~Khoury, Q.-Y. Zhou, and V.~Koltun.
\newblock Learning compact geometric features.
\newblock In {\em The IEEE International Conference on Computer Vision (ICCV)},
  Oct 2017.

\bibitem{korman2018latent}
S.~Korman and R.~Litman.
\newblock Latent ransac.
\newblock In {\em Proceedings of the IEEE Conference on Computer Vision and
  Pattern Recognition}, pages 6693--6702, 2018.

\bibitem{krizhevsky2012imagenet}
A.~Krizhevsky, I.~Sutskever, and G.~E. Hinton.
\newblock Imagenet classification with deep convolutional neural networks.
\newblock In {\em Advances in neural information processing systems}, pages
  1097--1105, 2012.

\bibitem{li20073d}
H.~Li and R.~Hartley.
\newblock The 3d-3d registration problem revisited.
\newblock In {\em Computer Vision, 2007. ICCV 2007. IEEE 11th International
  Conference on}, pages 1--8. IEEE, 2007.

\bibitem{lowe2004distinctive}
D.~G. Lowe.
\newblock Distinctive image features from scale-invariant keypoints.
\newblock {\em International journal of computer vision}, 60(2):91--110, 2004.

\bibitem{mellado2014super}
N.~Mellado, D.~Aiger, and N.~J. Mitra.
\newblock Super 4pcs fast global pointcloud registration via smart indexing.
\newblock In {\em Computer Graphics Forum}, volume~33, pages 205--215. Wiley
  Online Library, 2014.

\bibitem{mohamad2015super}
M.~Mohamad, M.~T. Ahmed, D.~Rappaport, and M.~Greenspan.
\newblock Super generalized 4pcs for 3d registration.
\newblock In {\em 3D Vision (3DV), 2015 International Conference on}, pages
  598--606. IEEE, 2015.

\bibitem{park2017colored}
J.~Park, Q.-Y. Zhou, and V.~Koltun.
\newblock Colored point cloud registration revisited.
\newblock In {\em Proceedings of the IEEE Conference on Computer Vision and
  Pattern Recognition}, pages 143--152, 2017.

\bibitem{pytorch}
A.~Paszke, S.~Gross, S.~Chintala, G.~Chanan, E.~Yang, Z.~DeVito, Z.~Lin,
  A.~Desmaison, L.~Antiga, and A.~Lerer.
\newblock Automatic differentiation in pytorch.
\newblock In {\em NIPS-Workshops}, 2017.

\bibitem{petrelli2011repeatability}
A.~Petrelli and L.~Di~Stefano.
\newblock On the repeatability of the local reference frame for partial shape
  matching.
\newblock In {\em Computer Vision (ICCV), 2011 IEEE International Conference
  on}, pages 2244--2251. IEEE, 2011.

\bibitem{raguram2013usac}
R.~Raguram, O.~Chum, M.~Pollefeys, J.~Matas, and J.-M. Frahm.
\newblock Usac: a universal framework for random sample consensus.
\newblock {\em IEEE Trans. Pattern Anal. Mach. Intell.}, 35(8):2022--2038,
  2013.

\bibitem{rusu2009fast}
R.~B. Rusu, N.~Blodow, and M.~Beetz.
\newblock Fast point feature histograms (fpfh) for 3d registration.
\newblock In {\em Robotics and Automation, 2009. ICRA'09. IEEE International
  Conference on}, pages 3212--3217. IEEE, 2009.

\bibitem{salti2014shot}
S.~Salti, F.~Tombari, and L.~Di~Stefano.
\newblock Shot: Unique signatures of histograms for surface and texture
  description.
\newblock {\em Computer Vision and Image Understanding}, 125:251--264, 2014.

\bibitem{shotton2013scene}
J.~Shotton, B.~Glocker, C.~Zach, S.~Izadi, A.~Criminisi, and A.~Fitzgibbon.
\newblock Scene coordinate regression forests for camera relocalization in
  rgb-d images.
\newblock In {\em Proceedings of the IEEE Conference on Computer Vision and
  Pattern Recognition}, pages 2930--2937, 2013.

\bibitem{toldo2010global}
R.~Toldo, A.~Beinat, and F.~Crosilla.
\newblock Global registration of multiple point clouds embedding the
  generalized procrustes analysis into an icp framework.
\newblock In {\em 3DPVT 2010 Conference}, 2010.

\bibitem{tombari2010unique}
F.~Tombari, S.~Salti, and L.~Di~Stefano.
\newblock Unique shape context for 3d data description.
\newblock In {\em Proceedings of the ACM workshop on 3D object retrieval},
  pages 57--62. ACM, 2010.

\bibitem{vidal20186d}
J.~Vidal, C.-Y. Lin, and R.~Mart{\'\i}.
\newblock 6d pose estimation using an improved method based on point pair
  features.
\newblock In {\em 2018 4th International Conference on Control, Automation and
  Robotics (ICCAR)}, pages 405--409. IEEE, 2018.

\bibitem{vongkulbhisal2018inverse}
J.~Vongkulbhisal, B.~I. Ugalde, F.~De~la Torre, and J.~P. Costeira.
\newblock Inverse composition discriminative optimization for point cloud
  registration.
\newblock In {\em Proceedings of the IEEE Conference on Computer Vision and
  Pattern Recognition}, pages 2993--3001, 2018.

\bibitem{worrall2018}
D.~Worrall and G.~Brostow.
\newblock Cubenet: Equivariance to 3d rotation and translation.
\newblock In V.~Ferrari, M.~Hebert, C.~Sminchisescu, and Y.~Weiss, editors,
  {\em Computer Vision -- ECCV 2018}, 2018.

\bibitem{xiao2013sun3d}
J.~Xiao, A.~Owens, and A.~Torralba.
\newblock Sun3d: A database of big spaces reconstructed using sfm and object
  labels.
\newblock In {\em Proceedings of the IEEE International Conference on Computer
  Vision}, pages 1625--1632, 2013.

\bibitem{yang2013go}
J.~Yang, H.~Li, and Y.~Jia.
\newblock Go-icp: Solving 3d registration efficiently and globally optimally.
\newblock In {\em Proceedings of the IEEE International Conference on Computer
  Vision}, pages 1457--1464, 2013.

\bibitem{Yang_2018_CVPR}
Y.~Yang, C.~Feng, Y.~Shen, and D.~Tian.
\newblock Foldingnet: Point cloud auto-encoder via deep grid deformation.
\newblock In {\em The IEEE Conference on Computer Vision and Pattern
  Recognition (CVPR)}, June 2018.

\bibitem{yi2016lift}
K.~M. Yi, E.~Trulls, V.~Lepetit, and P.~Fua.
\newblock Lift: Learned invariant feature transform.
\newblock In {\em European Conference on Computer Vision}, pages 467--483.
  Springer, 2016.

\bibitem{zeng20163dmatch}
A.~Zeng, S.~Song, M.~Nie{\ss}ner, M.~Fisher, J.~Xiao, and T.~Funkhouser.
\newblock 3dmatch: Learning local geometric descriptors from rgb-d
  reconstructions.
\newblock In {\em CVPR}, 2017.

\bibitem{zhao20193d}
Y.~Zhao, T.~Birdal, H.~Deng, and F.~Tombari.
\newblock 3d point-capsule networks.
\newblock In {\em IEEE Conf. on Computer Vision and Pattern Recognition
  (CVPR)}, June 2019.

\bibitem{zhou2016fast}
Q.-Y. Zhou, J.~Park, and V.~Koltun.
\newblock Fast global registration.
\newblock In {\em European Conference on Computer Vision}, pages 766--782.
  Springer, 2016.

\end{thebibliography}
% }

\end{document}